\documentclass[preprint,12pt]{elsarticle}
\usepackage{cmap}
\usepackage[T1]{fontenc}
\usepackage{lmodern}

\usepackage[cmex10]{amsmath}
\usepackage{amssymb}
\usepackage{amsfonts}
\usepackage{graphicx}
\graphicspath{{Figures/}}
\DeclareGraphicsExtensions{.pdf,.jpeg,.png}

\usepackage[section]{placeins}
\usepackage{booktabs}
\usepackage{array}
\usepackage{multirow}
\usepackage[caption=false,font=footnotesize]{subfig}
\usepackage{url}
\usepackage{xcolor}
\usepackage{hyperref}

\journal{arXiv preprint}

\begin{document}
\sloppy

\begin{frontmatter}

\title{Well-Log Reconstruction with Pre-trained Geological Knowledge: A Knowledge-Conditioned Large-Language-Model Generative Agent}

\author[cup]{Yongan Zhang}
\ead{zhangyongan.polyu@arctura-labs.com}

\author[cup]{Yitao Yu}
\ead{yuyitao.upc@arctura-labs.com}

\author[cup]{Yizheng Li}
\ead{liyizheng.polyu@arctura-labs.com}

\author[cup]{An Gong}
\ead{gongan.upc@arctura-labs.com}

\author[cup]{Youzhuang Sun\corref{cor1}}
\ead{sunyouzhuang.upc@arctura-labs.com}
\cortext[cor1]{Corresponding author.}

\affiliation[cup]{organization={College of Computer Science and Technology, China University of Petroleum (East China)},
            city={Qingdao},
            postcode={266580},
            country={China}}

\begin{abstract}
Petroleum and natural gas exploration remains fundamental to global energy security, and geophysical well logging provides the primary subsurface measurements required to evaluate deep hydrocarbon reservoirs; however, key petrophysical logs frequently suffer from extended interval absence in complex drilling environments. Existing deep learning reconstruction models rely solely on single-well numerical regression and lack macro-scale geological prior knowledge to constrain ill-posed inverse mappings, leaving open how domain geological priors can be systematically incorporated into continuous waveform synthesis to resolve multi-solution ambiguity across heterogeneous formations. To address this gap, we introduce pre-trained large language models (LLMs) as a source of encyclopedic geological priors and propose a knowledge-conditioned generative artificial intelligence (Generative AI) agent (\textbf{GeoLog-Agent}) coupled via a universal harness (\textbf{Geo-Harness}) that translates LLM stratigraphic reasoning into formation-anchored petrophysical trajectories, feature-wise linear modulation, and conditional adversarial constraints. Comprehensive comparative, ablation, and sensitivity experiments were conducted on the Norwegian North Sea benchmark dataset. In comparative experiments against nine mainstream deep learning baselines, \textbf{GeoLog-Agent} achieves a mean $R^2$ of $0.9307$ and a peak $R^2$ of $0.9713$, outperforming the nine-baseline average ($0.8387$) by $0.0920$ in $R^2$ ($30.97\%$ average RMSE reduction) and surpassing the strongest baseline by $0.0274$ to $0.0498$ in $R^2$ ($16.25\%$ to $24.05\%$ RMSE reduction). Ablation experiments confirm both the individual effectiveness and cross-module synergy of the geological prior reasoning, multi-scale dilated generation, and conditional adversarial modules, while sensitivity experiments show that transferring this coupling mechanism to six conventional deep learning models consistently improves their $R^2$ by $0.0162$ to $0.0789$ with high cross-well stability. This study provides reliable petrophysical constraints for deep hydrocarbon reservoir characterization in unlogged wellbore intervals and establishes a generalizable scientific paradigm that couples foundation-model domain reasoning with deep generative networks to solve ill-posed inverse problems across geoscience and engineering.
\end{abstract}

\begin{keyword}
Well-log reconstruction \sep Large language model \sep Generative artificial intelligence (Generative AI) \sep Geological prior knowledge \sep Knowledge-guided generative paradigm \sep Petroleum exploration
\end{keyword}

\end{frontmatter}

\section{Introduction}
\label{sec:intro}

Petroleum and natural gas resources remain the cornerstone of global energy supply and industrial stability, and discovering and developing deep subsurface hydrocarbon reservoirs is a central task of geoenergy engineering \cite{darling2005well,lai2022geophysical}. Because deep oil and gas formations lie thousands of meters beneath the Earth's surface and cannot be observed directly, geophysical well logging serves as the primary subsurface sensing technology for recording continuous vertical rock-property profiles along drilled wellbores. Among standard wireline measurements, porosity- and elasticity-sensitive curves such as bulk density (DEN) and compressional acoustic transit time (AC) provide essential constraints for identifying hydrocarbon-bearing reservoirs and calibrating seismic profiles. In field operations across offshore and deep onshore basins, however, wireline logs frequently suffer from extended interval absence caused by borehole wall collapse, tool sticking in swelling clays, radioactive source restrictions, or cost-saving omissions in older appraisal wells \cite{gama2025imputation}. Because incomplete well-log suites interrupt reservoir evaluation across entire oilfields, accurately reconstructing missing well-log curves from routinely acquired conventional logs such as gamma ray (GR), caliper (CALI), spontaneous potential (SP), neutron porosity (NPHI), and medium and deep resistivity (RMED and RDEP) has become a foundational computational problem in petroleum exploration \cite{feng2021imputation}.

Over the past decade, data-driven well-log reconstruction has progressed from pointwise empirical regressions and shallow decision trees \cite{mo2015well,feng2021imputation} toward deep sequence modeling and generative artificial intelligence (Generative AI) architectures capable of capturing vertical stratigraphic dependencies. Recurrent neural networks, including Long Short-Term Memory (LSTM) networks, Bidirectional LSTM (BiLSTM) models, and bidirectional recurrent imputation frameworks \cite{cao2018brits,pham2020missing,zeng2020attention,yang2023deep,zhang2024well}, model sequential transitions along the wellbore but suffer from hidden-state drift when missing intervals extend across hundreds of meters. To expand the vertical receptive field, Transformer-based sequence architectures such as vanilla Transformer \cite{ashish2017attention,lin2023missing}, Informer \cite{zhou2021informer}, Autoformer \cite{wu2021autoformer}, PatchTST \cite{nie2023patchtst}, and TimesNet \cite{wu2023timesnet}, alongside linear decomposition baselines such as DLinear \cite{zeng2023dlinear}, have been adapted to model multi-scale wireline correlations. Concurrently, deep Generative AI models, particularly Generative Adversarial Networks (GANs) and conditional GANs, have been introduced to mitigate the over-smoothing inherent in pure norm-minimization objectives and synthesize realistic high-frequency wireline textures \cite{mirza2014conditional,qu2024novel,al2024utilizing,al2025well,shi2025missing}.

Despite these architectural advances, existing deep learning reconstruction models remain confined to single-well numerical regression and lack macro-scale geological prior knowledge to constrain ill-posed inverse mappings across heterogeneous formations. Across sedimentary basins worldwide, inferring target petrophysical logs from conventional auxiliary curves is fundamentally a non-bijective, multi-solution inverse problem because distinct geological formations and depositional environments frequently produce overlapping conventional wireline responses while following opposite density and velocity laws. A representative field manifestation occurs in the Norwegian North Sea South Viking Graben \cite{bormann2020force}, where Paleocene and Eocene volcaniclastic tuff beds in the Balder Formation exhibit elevated gamma-ray and neutron readings that closely resemble ordinary marine shales yet possess a stiffer mineral framework, and where low-gamma-ray intervals can correspond either to unconsolidated sandstones in the Utsira and Frigg Formations or to dense pelagic carbonates in the Shetland Group. Relying solely on local wireline numbers without macro-scale geological priors, purely data-driven deep models cannot separate these competing physical regimes and inevitably average across incompatible rock formations. How to systematically incorporate macro-scale geological prior knowledge into continuous waveform generation to constrain ill-posed inverse mappings and eliminate cross-formation multi-solution ambiguity therefore constitutes a critical scientific gap.

Leaving this gap unresolved carries direct operational consequences for hydrocarbon exploration and reservoir engineering. Subsurface workflows rely on reconstructed well logs to calibrate well-to-seismic ties, estimate reservoir storage capacity, and delineate productive hydrocarbon pay zones across unlogged wellbore intervals. When purely numerical reconstructors smooth out formation boundaries or misidentify lithological regimes due to geological context blindness, the resulting errors propagate directly into static and dynamic reservoir models, causing engineers to misjudge caprock integrity or overlook productive pay intervals. Acquiring such macro-scale geological priors traditionally required manual expert interpretation that cannot scale across automated deep learning pipelines; however, recent breakthroughs in pre-trained large language models (LLMs) offer a transformative solution to supply this missing prior knowledge by encoding encyclopedic stratigraphic and rock-physics laws distilled from massive geoscientific corpora \cite{deepseek2024v3,deepseek2025r1}. Introducing pre-trained LLMs to provide macro-scale geological priors and coupling them with deep generative networks enables blind-well reconstruction to dynamically adapt its physical mapping across changing geological formations, ensuring both basin-scale stratigraphic consistency and wellbore-scale waveform fidelity.

\begin{figure*}[!t]
\centering
\includegraphics[width=\textwidth]{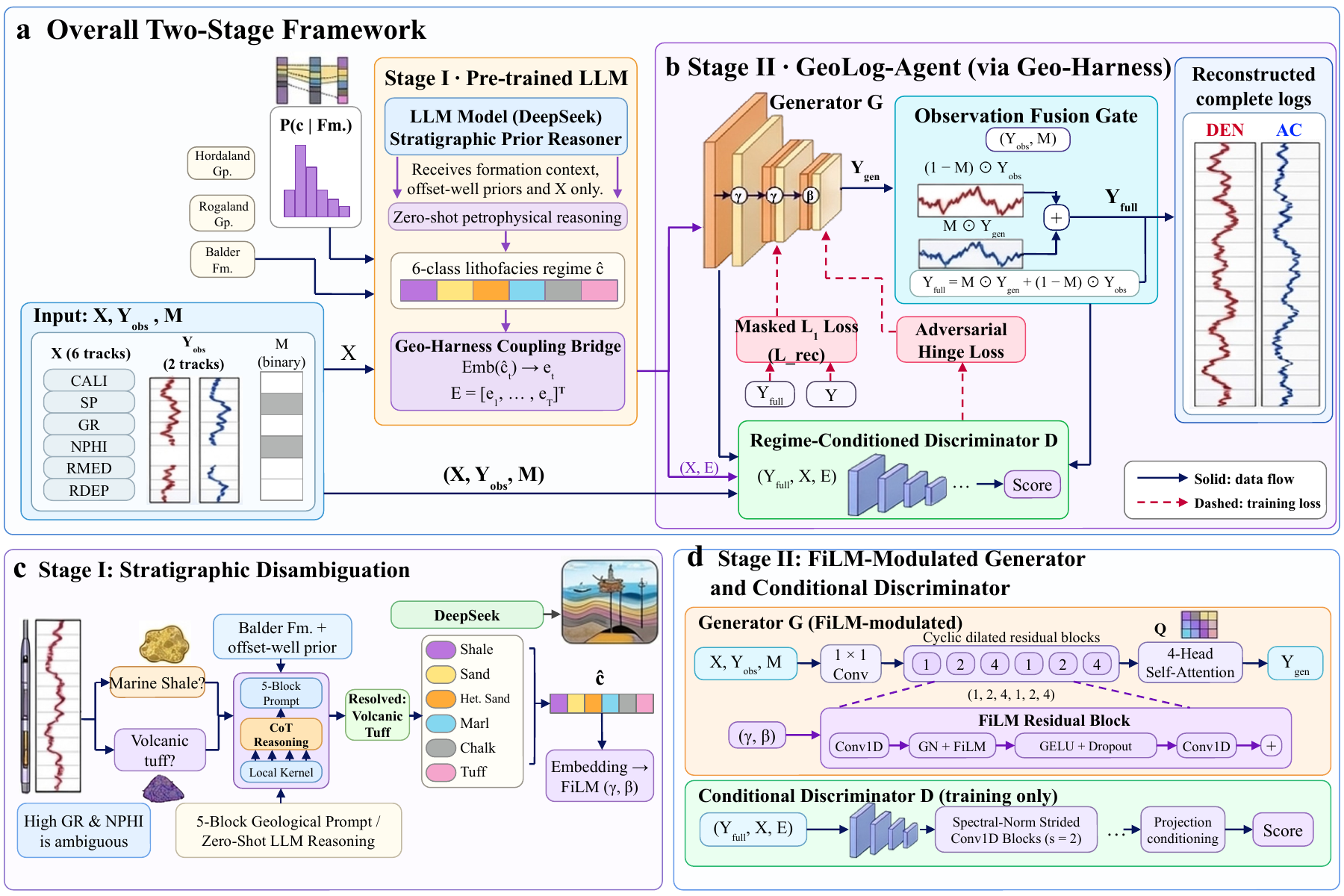}
\caption{Architecture of \textbf{GeoLog-Agent} and the \textbf{Geo-Harness} coupling bridge: (\textbf{a}) two-stage collaborative workflow, (\textbf{b}) Stage~II knowledge-modulated adversarial training and observation fusion, (\textbf{c}) Stage~I pre-trained LLM geological prior reasoning, and (\textbf{d}) FiLM-modulated cyclic dilated generator $G$ and spectral-normalized discriminator $D$.}
\label{fig:framework}
\end{figure*}

Building on this insight, we introduce pre-trained LLMs as a macro-scale geological prior engine and propose a knowledge-conditioned generative AI agent (\textbf{GeoLog-Agent}) coupled through a universal, plug-and-play geological knowledge harness (\textbf{Geo-Harness}, Fig.~\ref{fig:framework}). In Stage~I, a pre-trained LLM geological reasoner (instantiated primarily with \texttt{DeepSeek-V3-0324} \cite{deepseek2024v3} and evaluated alongside \texttt{DeepSeek-R1-0528} chain-of-thought reasoning \cite{deepseek2025r1}) synthesizes formation-top metadata, offset-well lithofacies priors, and depth-windowed conventional logs to infer a depth-aligned six-class petrophysical regime sequence across blind test wells without target-well retraining. In Stage~II, \textbf{Geo-Harness} bridges symbolic LLM reasoning and continuous deep generative synthesis by compiling formation-anchored rock-physics prior trajectories, modulating a multi-scale cyclic dilated residual generator via Feature-wise Linear Modulation (FiLM) \cite{perez2018film,yu2015multi}, and constraining a spectral-normalized conditional discriminator \cite{miyato2018spectral} with boundary-continuous observation fusion. Beyond a single model architecture, this study investigates and validates a generalizable scientific paradigm in which pre-trained foundation models guide deep sequence and generative models (LSTM, BiLSTM, Transformer, Informer, PatchTST, and GAN) across non-stationary subsurface regimes, as confirmed by comprehensive comparative, ablation, and sensitivity experiments on the open-source Norwegian North Sea benchmark \cite{bormann2020force}.

\section{Methodology}
\label{sec:method}

\subsection{Problem Formulation}

Let a multi-curve wireline logging sequence along a wellbore of length $T$ depth samples be represented by an auxiliary input matrix $\mathbf{X} = [\mathbf{x}_1, \mathbf{x}_2, \dots, \mathbf{x}_T]^\top \in \mathbb{R}^{T \times d_x}$ and a target porosity-sensitive log matrix $\mathbf{Y} = [\mathbf{y}_1, \mathbf{y}_2, \dots, \mathbf{y}_T]^\top \in \mathbb{R}^{T \times d_y}$. In the Norwegian North Sea dataset examined in this work, the auxiliary log suite comprises $d_x = 6$ conventional wireline curves (GR, CALI, SP, NPHI, RMED, and RDEP), while the target matrix comprises $d_y = 2$ curves corresponding to bulk density (DEN, $\text{g/cm}^3$) and compressional acoustic transit time (AC, $\mu\text{s/m}$). To account for cross-well spontaneous potential baseline shifts and orders-of-magnitude resistivity variations across hydrocarbon and brine intervals, raw resistivity curves are transformed into logarithmic space via $\tilde{R} = \log_{10}(\mathrm{clip}(R, 0.05, 2000.0))$ and raw SP is centered relative to the per-well shale median baseline $\tilde{\mathrm{SP}}_t = \mathrm{SP}_t - \mathrm{median}(\mathrm{SP})$ prior to standardization.

Target curve absence is encoded by a binary missingness mask $\mathbf{M} \in \{0, 1\}^{T \times d_y}$, where $M_{t,j} = 1$ indicates that target curve $j$ at depth index $t$ is unrecorded and requires reconstruction, whereas $M_{t,j} = 0$ indicates a valid borehole measurement. The partially observed target tensor available at inference time is defined by element-wise masking as
\begin{equation}
\mathbf{Y}_{\mathrm{obs}} = (\mathbf{1} - \mathbf{M}) \odot \mathbf{Y},
\end{equation}
where $\odot$ denotes the Hadamard product. Under extended missing intervals, local interpolation fails because observational anchors are absent across hundreds to thousands of consecutive depth steps.

More critically, the conditional distribution $p(\mathbf{y}_t \mid \mathbf{x}_t)$ is multimodal across distinct geological regimes $c_t \in \{0, 1, \dots, K-1\}$. Under Wyllie time-average and Gardner density--velocity relations, bulk density $\rho_b$ and compressional transit time $\Delta t$ depend convexly on both mineral matrix properties ($\rho_{\mathrm{ma}}(c_t), \Delta t_{\mathrm{ma}}(c_t)$) and pore-fluid saturation. Consequently, two depth intervals belonging to different lithofacies regimes $c_t = k_1$ and $c_t = k_2$ can exhibit nearly identical conventional log vectors $\mathbf{x}_t$ while possessing opposite density and sonic deviations. Conditioning the generative mapping on a knowledge-inferred regime sequence $\hat{\mathbf{c}} = [\hat{c}_1, \hat{c}_2, \dots, \hat{c}_T]^\top \in \{0, 1, \dots, K-1\}^T$ decomposes the ill-posed marginal mapping into unimodal regime-specific distributions $p(\mathbf{y}_t \mid \mathbf{x}_t, \hat{c}_t)$, thereby compressing conditional entropy:
\begin{equation}
\mathcal{H}(\mathbf{Y} \mid \mathbf{X}, \hat{\mathbf{c}}) \le \mathcal{H}(\mathbf{Y} \mid \mathbf{X}).
\end{equation}
As illustrated in Fig.~\ref{fig:framework}, the proposed \textbf{GeoLog-Agent} operates through a two-stage collaborative paradigm in which a pre-trained LLM geological reasoner (Stage~I) infers $\hat{\mathbf{c}}$ from regional stratigraphic context and conventional logs, and the \textbf{Geo-Harness} coupling bridge (Stage~II) modulates the downstream generative network to synthesize the complete target profile $\hat{\mathbf{Y}}_{\mathrm{full}}$.

\subsection{Pre-trained Geological Prior Reasoning}

When DEN and AC are missing over extended intervals in a blind well, a purely numerical classifier $f_{\mathrm{num}}(\mathbf{x}_t)$ trained solely on the six conventional logs suffers from severe regime confusion because the primary lithology-discriminating crossplots (such as the neutron--density $\mathrm{NPHI}$--$\mathrm{DEN}$ and sonic--density $\mathrm{AC}$--$\mathrm{DEN}$ crossplots) are unavailable. However, standard mud-logging records and regional seismic-stratigraphic markers routinely provide the geological group ($\mathrm{Grp}_t$) and formation ($\mathrm{Fm}_t$) tops penetrated by the wellbore, alongside statistical petrophysical summaries from offset wells in the same basin.

To exploit this non-numerical geological knowledge, Stage~I employs a pre-trained large language model ($\mathcal{F}_{\mathrm{LLM}}$, instantiated with \texttt{DeepSeek-V3-0324} \cite{deepseek2024v3} as the default Stage~I reasoner and \texttt{DeepSeek-R1-0528} \cite{deepseek2025r1} as a chain-of-thought sensitivity reference) as a zero-shot geological prior reasoner. The wellbore is partitioned into contiguous stratigraphic segments $S_m = [t_m^{\mathrm{start}}, t_m^{\mathrm{end}}]$ bounded primarily by lithostratigraphic formation tops, with thick formations subdivided at maximum wireline $L_2$-gradient change points to maintain a target resolution of $\sim 90$ segments per well (mean window length $\sim 20\text{ m} \approx 132$ depth samples at $0.152\text{ m}$ spacing, minimum spacing $15$ samples). For each segment $S_m$, a structured geological prompt $\mathcal{P}_m$ is constructed by combining three information sources:
\begin{equation}
\mathcal{P}_m = \Big\{ \mathcal{I}_{\mathrm{basin}}(\mathrm{Well}, z_m, \mathrm{Grp}_m, \mathrm{Fm}_m), \; \mathcal{S}_{\mathrm{offset}}(\mathrm{Fm}_m), \; \bar{\mathbf{x}}_{S_m} \Big\},
\end{equation}
where $\mathcal{I}_{\mathrm{basin}}$ encodes the regional basin identity (Norwegian North Sea South Viking Graben), measured depth interval $z_m$, and stratigraphic unit names; $\mathcal{S}_{\mathrm{offset}}(\mathrm{Fm}_m)$ provides the empirical lithofacies regime prior distribution and formation-specific log statistics ($\mu_{k,\mathrm{Fm}}, \sigma_{k,\mathrm{Fm}}$) compiled strictly from the eight offset training wells; and $\bar{\mathbf{x}}_{S_m}$ summarizes the mean, standard deviation, and vertical gradient of the six available conventional logs ($\mathrm{GR}, \mathrm{CALI}, \tilde{\mathrm{SP}}, \mathrm{NPHI}, \tilde{\mathrm{RMED}}, \tilde{\mathrm{RDEP}}$) within segment $S_m$.

Each prompt $\mathcal{P}_m$ uses deterministic JSON decoding ($\mathrm{temperature} = 0.0$) to output a normalized six-class regime probability vector $\mathbf{p}_{\mathrm{LLM}}(k \mid \mathcal{P}_m)$ ($k \in \{0,\dots,5\}$) and a concise petrophysical rationale. Using its pre-trained geological knowledge base and structured chain-of-thought reasoning, the LLM evaluates whether the observed conventional log signature $\bar{\mathbf{x}}_{S_m}$ matches the depositional lithofacies expected within formation $\mathrm{Fm}_m$, such as recognizing that high-GR, high-NPHI intervals inside the Eocene Balder Formation represent submarine volcanic tuff beds rather than ordinary Hordaland marine shales, or distinguishing low-GR pelagic chalk in the Tor Formation from clean siliciclastic sands in the Heimdal and Frigg Formations.

\subsection{Knowledge-Modulated Generative Architecture}

A central methodological question in coupling pre-trained foundation models with deep domain models is how a symbolic, segment-level reasoner (Stage~I) and a continuous, sample-level neural generator (Stage~II) reinforce each other. Standalone LLMs possess broad encyclopedic knowledge of regional stratigraphy and rock physics but cannot directly output thousands of continuous floating-point wireline samples at $0.152\text{ m}$ vertical resolution without quantization drift. Conversely, deep sequence and generative models (such as CNNs, LSTMs, Transformers, and GANs) excel at capturing high-frequency local waveform correlations ($\partial \mathbf{x}_t / \partial z$) but lack macroscopic awareness of formation identity when target logs are absent.

To bridge these two complementary paradigms, we design the \textbf{Geological Knowledge Harness} (\textbf{Geo-Harness}), a universal knowledge-to-tensor coupling module that translates Stage~I LLM reasoning into three continuous conditioning signals for Stage~II:
\begin{enumerate}
  \item \textbf{Bayesian Depth-Local Posterior Alignment:} Combining the segment-level LLM posterior probabilities $\mathbf{p}_{\mathrm{LLM}}(k \mid \mathcal{P}_{m(t)})$ (smoothed across segment boundaries via a zero-phase 1D Gaussian kernel with $\sigma = 3$ samples $= 0.456\text{ m}$) with depth-local Gaussian kernel likelihoods $p_{\mathrm{local}}(\mathbf{x}_t \mid k, \mathrm{Fm}_t)$ over the six conventional logs yields the depth-aligned regime sequence $\hat{\mathbf{c}} = [\hat{c}_1, \dots, \hat{c}_T]^\top$:
  \begin{equation}
  \hat{c}_t = \arg\max_{k \in \{0,\dots,K-1\}} \Big( \log p_{\mathrm{LLM}}(k \mid \mathcal{P}_{m(t)}) + \log p_{\mathrm{local}}(\mathbf{x}_t \mid k, \mathrm{Fm}_t) \Big).
  \label{eq:posterior_fusion}
  \end{equation}
  \item \textbf{Formation-Regime Petrophysical Anchor Trajectories:} For each depth step $t$ with inferred regime $\hat{c}_t = k$ inside formation $\mathrm{Fm}_t$, \textbf{Geo-Harness} retrieves the offset-well rock-physics moments $\boldsymbol{\mu}_{\mathrm{Fm}_t, k}, \boldsymbol{\sigma}_{\mathrm{Fm}_t, k} \in \mathbb{R}^{d_y}$ and constructs multi-scale cross-channel interaction features $\mathbf{H}_{\mathrm{harness}}(t)$ (coupling the smoothed regime posteriors, rolling depth-window statistics over $w \in \{5, 15, 45, 101\}$ samples, and non-parametric rock-physics prior estimators $\boldsymbol{\mu}_{\mathrm{prior}}(t)$ fitted strictly on the eight offset training wells $\mathcal{D}_{\mathrm{train}}$ with zero target-well leakage).
  \item \textbf{Learnable Regime Embedding and FiLM Affine Modulation:} A learnable embedding table $\mathrm{Emb}(\cdot)$ projects each discrete regime index $\hat{c}_t$ and its prior anchor $\mathbf{H}_{\mathrm{harness}}(t)$ into a continuous conditioning vector $\mathbf{e}_t \in \mathbb{R}^{d_e}$ ($d_e = 32$):
  \begin{equation}
  \mathbf{e}_t = \mathrm{Emb}(\hat{c}_t) + \mathbf{W}_h \mathbf{H}_{\mathrm{harness}}(t), \quad \mathbf{E} = [\mathbf{e}_1, \mathbf{e}_2, \dots, \mathbf{e}_T]^\top \in \mathbb{R}^{T \times d_e}.
  \end{equation}
\end{enumerate}

Concatenating the auxiliary logs $\mathbf{X}$, the masked observations $\mathbf{Y}_{\mathrm{obs}}$, the binary mask $\mathbf{M}$, and the \textbf{Geo-Harness} conditioning representation $\mathbf{E}$ along the channel axis forms the composite generator input tensor $\mathbf{Z}_{\mathrm{in}} \in \mathbb{R}^{T \times (d_x + 2d_y + d_e)}$:
\begin{equation}
\mathbf{Z}_{\mathrm{in}} = [\mathbf{X}; \, \mathbf{Y}_{\mathrm{obs}}; \, \mathbf{M}; \, \mathbf{E}].
\end{equation}
Because \textbf{Geo-Harness} exposes both the standardized conditioning tensor $\mathbf{E}$ and the formation-anchored petrophysical prior trajectory $\boldsymbol{\mu}_{\mathrm{prior}}(t)$, it can either drive our full FiLM-modulated adversarial generator (\textbf{GeoLog-Agent}) or wrap any off-the-shelf sequence backbone $\mathcal{B}_\theta$ (e.g., LSTM, BiLSTM, Transformer, Informer, PatchTST, or GAN) in a plug-and-play manner via regime-conditioned prior-guided calibration ($\hat{\mathbf{y}}_{\mathrm{harness}}(t) = (1 - \alpha_{\hat{c}_t})\,\mathcal{B}_\theta(\mathbf{X})_t + \alpha_{\hat{c}_t}\,\boldsymbol{\mu}_{\mathrm{prior}}(\mathbf{x}_t, \hat{c}_t, \mathrm{Fm}_t)$, calibrated strictly on $\mathcal{D}_{\mathrm{train}}$).

In \textbf{GeoLog-Agent} (Fig.~\ref{fig:framework}d), a pointwise $1 \times 1$ convolutional stem projects $\mathbf{Z}_{\mathrm{in}}$ into a latent feature sequence $\mathbf{H}^{(0)} \in \mathbb{R}^{T \times d_h}$ ($d_h = 128$). To capture both centimeter-scale thin-bed interfaces and decameter-scale sedimentary cycles without spatial downsampling, the generator backbone stacks $L = 6$ residual blocks with cyclic dilation factors $r_l \in \{1, 2, 4, 1, 2, 4\}$ \cite{yu2015multi,he2016deep}. Matching the block pipeline in Fig.~\ref{fig:framework}d, each residual block $l \in \{1, \dots, L\}$ applies dilated 1D convolution ($\mathrm{Conv1D}_{r_l}$), Group Normalization ($\mathrm{GN}$) with Feature-wise Linear Modulation (FiLM) \cite{perez2018film} driven by the \textbf{Geo-Harness} embedding $\mathbf{E}$, gated activation ($\tanh(\cdot) \odot \sigma(\cdot)$), and a $1\times 1$ residual projection ($\mathrm{Conv1D}_{k=1}$):
\begin{equation}
\tilde{\mathbf{U}}^{(l)} = \big( \mathbf{1} + \boldsymbol{\gamma}^{(l)}(\mathbf{E}) \big) \odot \mathrm{GN}\!\big( \mathrm{Conv1D}_{r_l}(\mathbf{H}^{(l-1)}) \big) + \boldsymbol{\beta}^{(l)}(\mathbf{E}),
\end{equation}
\begin{equation}
\mathbf{A}^{(l)} = \tanh\!\big( \tilde{\mathbf{U}}_{\mathrm{f}}^{(l)} \big) \odot \sigma\!\big( \tilde{\mathbf{U}}_{\mathrm{g}}^{(l)} \big),
\end{equation}
\begin{equation}
\mathbf{H}^{(l)} = \mathbf{H}^{(l-1)} + \mathrm{Dropout}\Big( \mathrm{Conv1D}_{k=1}\big( \mathbf{A}^{(l)} \big) \Big),
\end{equation}
where $\tilde{\mathbf{U}}^{(l)} \in \mathbb{R}^{T \times 2d_h}$ is split along the channel axis into filter and gate halves $\tilde{\mathbf{U}}_{\mathrm{f}}^{(l)}, \tilde{\mathbf{U}}_{\mathrm{g}}^{(l)} \in \mathbb{R}^{T \times d_h}$ (with $\mathrm{Conv1D}_{r_l}$ projecting $d_h \to 2d_h$), and $\boldsymbol{\gamma}^{(l)}(\mathbf{E}), \boldsymbol{\beta}^{(l)}(\mathbf{E})$ are regime-specific affine scale and shift parameters produced by linear projections of $\mathbf{E}$. This FiLM mechanism dynamically recalibrates the slope and baseline of the convolutional filters whenever the wellbore crosses a stratigraphic boundary.

To further capture long-range cross-formation dependencies across extended missing intervals, the dilated backbone output $\mathbf{H}^{(L)}$ is processed by a two-layer multi-head self-attention context encoder ($\mathrm{CtxAttn}$) with four attention heads ($d_{\mathrm{head}} = 32$), followed by a knowledge-conditioned residual skip projection anchored around $\boldsymbol{\mu}_{\mathrm{prior}}$:
\begin{equation}
\mathbf{H}_{\mathrm{ctx}} = \mathrm{CtxAttn}(\mathbf{H}^{(L)}),
\end{equation}
\begin{equation}
\hat{\mathbf{Y}}_{\mathrm{gen}} = \boldsymbol{\mu}_{\mathrm{prior}} + \mathrm{Conv1D}_{k=1}(\mathbf{H}_{\mathrm{ctx}}) + \mathrm{MLP}_{\mathrm{skip}}\big( [\mathbf{X}; \, \mathbf{E}] \big).
\end{equation}

\subsection{Adversarial Optimization Strategy}

Because measured wireline data in unmasked intervals ($M_{t,j} = 0$) represent physical ground truth, directly outputting $\hat{\mathbf{Y}}_{\mathrm{gen}}$ across the entire wellbore would introduce unnecessary approximation drift in observed shoulder intervals. \textbf{Geo-Harness} incorporates a boundary-continuous observation fusion gate that combines the generator prediction $\hat{\mathbf{Y}}_{\mathrm{gen}}$ with $\mathbf{Y}_{\mathrm{obs}}$ via:
\begin{equation}
\hat{\mathbf{Y}}_{\mathrm{full}} = \mathbf{M} \odot \hat{\mathbf{Y}}_{\mathrm{gen}} + (\mathbf{1} - \mathbf{M}) \odot \mathbf{Y}_{\mathrm{obs}},
\label{eq:fusion}
\end{equation}
with exponential shoulder-anchor calibration across the immediate boundary samples when observed shoulders exist ($M_{t,j}=0$). Equation~\eqref{eq:fusion} guarantees exact preservation of valid borehole measurements outside the missing region and eliminates step artifacts across the missing-to-observed boundary transition.

To prevent the over-smoothing produced by standalone $L_1$ or $L_2$ regression and to enforce regime-consistent petrophysical textures, a conditional projection discriminator $D_\phi(\hat{\mathbf{Y}}_{\mathrm{full}}, \mathbf{X}, \mathbf{E})$ evaluates the joint plausibility of the target curves, auxiliary logs, and stratigraphic regime embeddings (Fig.~\ref{fig:framework}b,d). Using four spectral-normalized strided 1D convolutional layers ($k=3, s=2$, hidden channels $64 \to 128 \to 128 \to 1$) with LeakyReLU activations ($\alpha = 0.2$) \cite{miyato2018spectral}, $D_\phi$ outputs a depth-wise realism score trajectory $D_\phi(\cdot, \mathbf{X}, \mathbf{E})$ optimized via the conditional hinge adversarial loss:
\begin{equation}
\mathcal{L}_D = \mathbb{E}\Big[ \max\big(0, \, 1 - D_\phi(\mathbf{Y}, \mathbf{X}, \mathbf{E})\big) \Big] + \mathbb{E}\Big[ \max\big(0, \, 1 + D_\phi(\hat{\mathbf{Y}}_{\mathrm{full}}, \mathbf{X}, \mathbf{E})\big) \Big],
\end{equation}
while the generator parameters $\theta$ are optimized by minimizing a composite objective combining the adversarial loss $\mathcal{L}_{\mathrm{adv}}$, a mask-normalized $L_1$ reconstruction loss $\mathcal{L}_{\mathrm{rec}}$, and a vertical gradient matching loss $\mathcal{L}_{\mathrm{grad}}$ evaluated on $\hat{\mathbf{Y}}_{\mathrm{full}}$ ($\lambda_{\mathrm{rec}} = 10.0$, $\lambda_{\mathrm{grad}} = 2.0$, $\varepsilon = 10^{-6}$; note that $\mathbf{M} \odot (\hat{\mathbf{Y}}_{\mathrm{full}} - \mathbf{Y}) \equiv \mathbf{M} \odot (\hat{\mathbf{Y}}_{\mathrm{gen}} - \mathbf{Y})$ by Equation~\eqref{eq:fusion}):
\begin{equation}
\mathcal{L}_{\mathrm{adv}} = -\mathbb{E}\Big[ D_\phi(\hat{\mathbf{Y}}_{\mathrm{full}}, \mathbf{X}, \mathbf{E}) \Big], \quad \mathcal{L}_{\mathrm{rec}} = \frac{\big\| \mathbf{M} \odot (\hat{\mathbf{Y}}_{\mathrm{full}} - \mathbf{Y}) \big\|_1}{\|\mathbf{M}\|_1 + \varepsilon},
\end{equation}
\begin{equation}
\mathcal{L}_{\mathrm{grad}} = \frac{\big\| \mathbf{M} \odot (\nabla_z \hat{\mathbf{Y}}_{\mathrm{full}} - \nabla_z \mathbf{Y}) \big\|_1}{\|\mathbf{M}\|_1 + \varepsilon}, \quad \mathcal{L}_G = \mathcal{L}_{\mathrm{adv}} + \lambda_{\mathrm{rec}} \mathcal{L}_{\mathrm{rec}} + \lambda_{\mathrm{grad}} \mathcal{L}_{\mathrm{grad}}.
\end{equation}

\section{Experimental Results}
\label{sec:results}

\subsection{Dataset Description}

Empirical evaluation is conducted on ten multi-formation exploration and appraisal wells from the open-source Norwegian North Sea FORCE 2020 benchmark released by the Norwegian Offshore Directorate \cite{bormann2020force}. The selected wells penetrate the Cenozoic, Cretaceous, and Triassic stratigraphic succession of the South Viking Graben (Quadrants 15 and 16, spanning the Sleipner, Volve, and Utsira High provinces), including the Nordland Group (Utsira Formation marine sands and claystones), the Hordaland Group, the Rogaland Group (Balder Formation submarine volcanic tuff, Sele Formation restricted marine shales, Lista Formation deep-marine shales, and Heimdal and Frigg Formations turbiditic fan sandstones), the Shetland Group (Ekofisk, Tor, Hod, and Blodoeks Formations pelagic chalks and marls), and the Cromer Knoll and Hegre Groups.

\begin{figure*}[!t]
\centering
\subfloat[Full-well M50 profile in Well 1 (15-9-15)]{\includegraphics[width=0.492\textwidth]{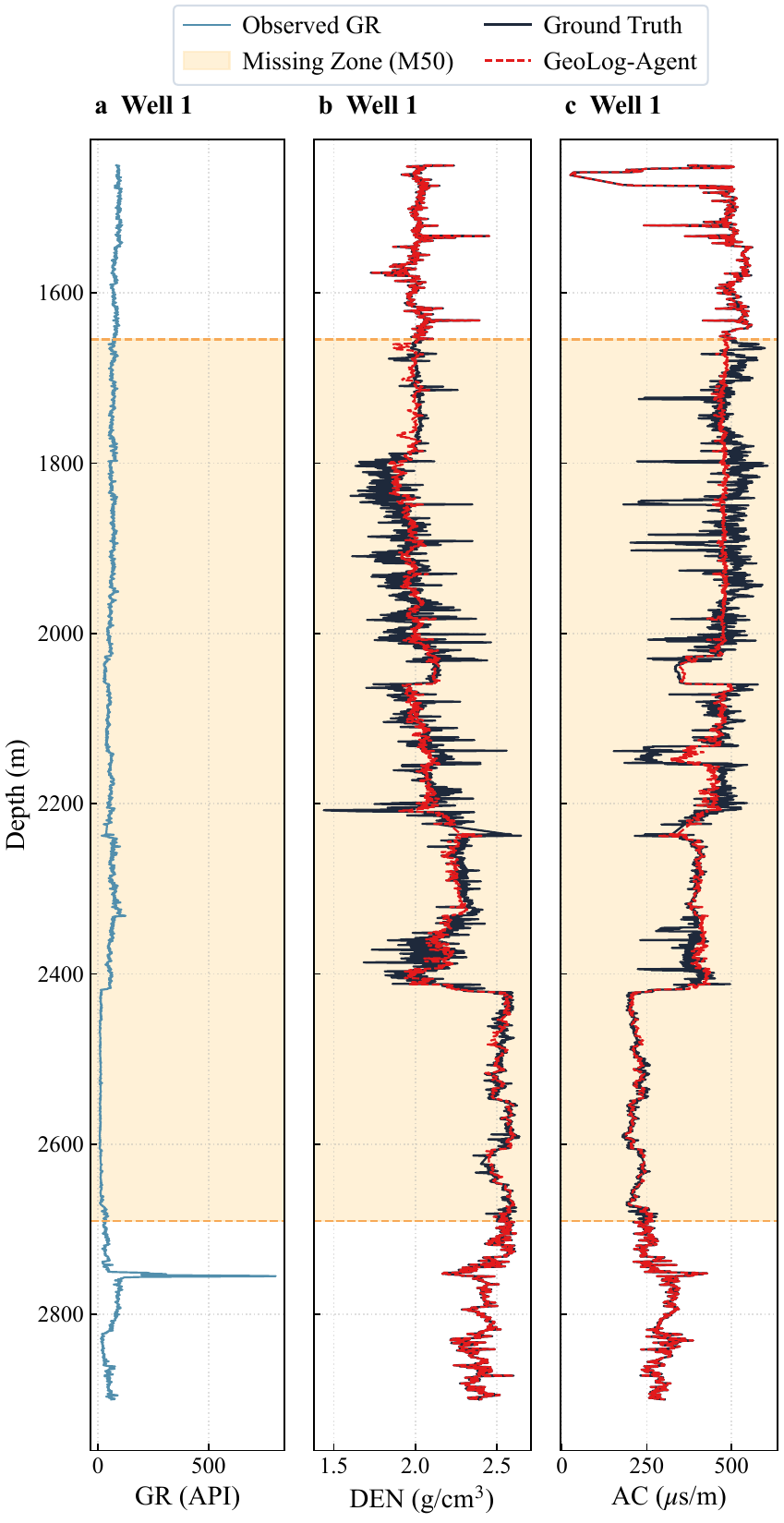}%
\label{fig:well1_overview}}
\hfil
\subfloat[Balder--Sele zoom in Well 1 and Well 2]{\includegraphics[width=0.492\textwidth]{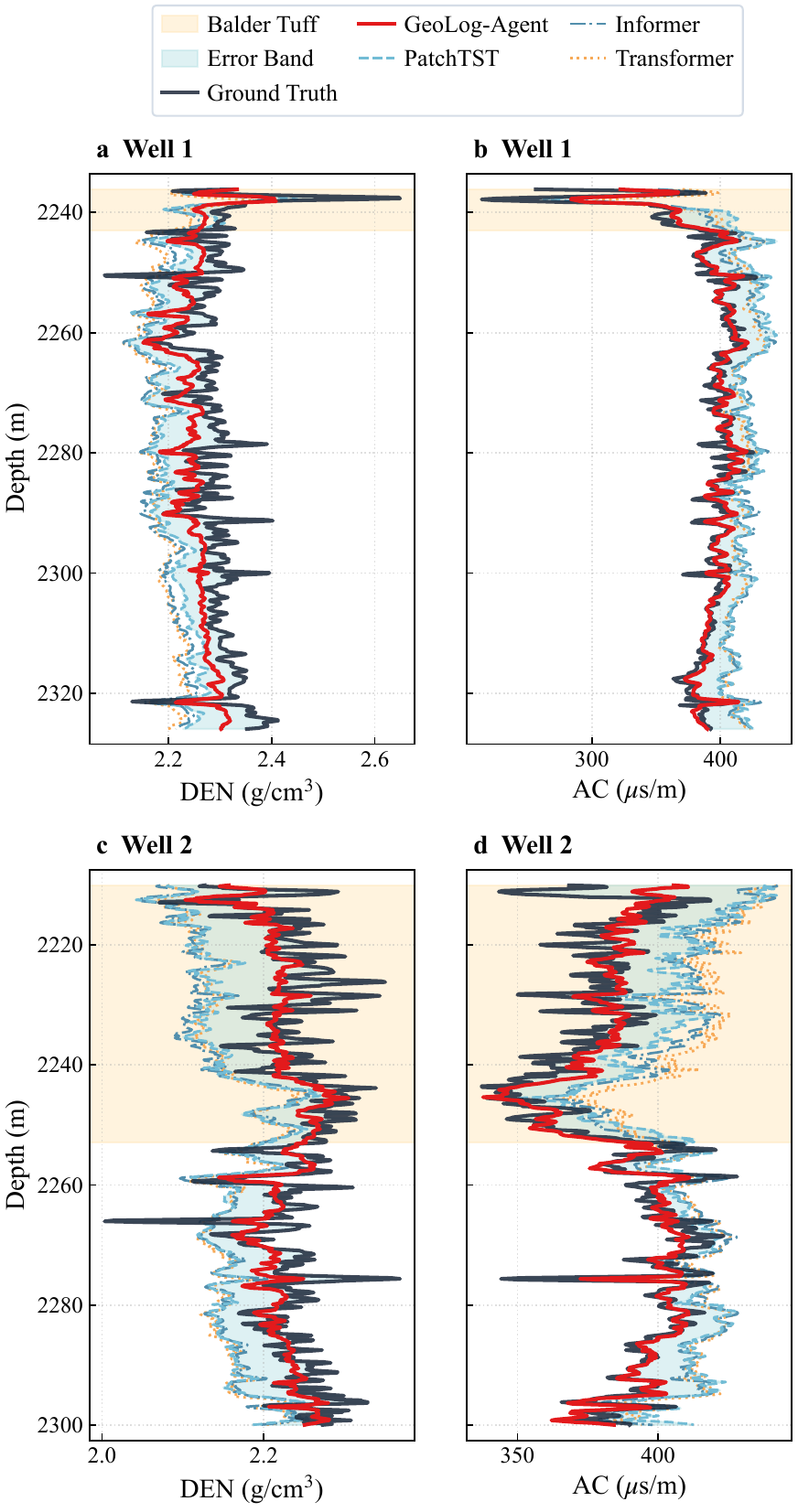}%
\label{fig:best3_comparison}}
\caption{Full-well reconstruction overview and local trajectory comparison: (\textbf{a}) observed GR, 50\% contiguous missing zone ($1654.6\text{--}2690.0\text{ m}$), and reconstructed DEN/AC profiles in Well 1; (\textbf{b}) local zoom across the Balder--Sele volcaniclastic transition in Well 1 ($2236\text{--}2326\text{ m}$, M50) and Well 2 ($2210\text{--}2300\text{ m}$, M100).}
\label{fig:dataset_overview}
\end{figure*}

To ensure strict cross-well blind generalization with zero wellbore overlap, eight wells (Wells 15-9-13, 16-10-1, 16-2-11-A, 16-2-16, 16-2-6, 16-4-1, 16-7-4, and 16-7-5, totaling 61,361 valid depth samples at a vertical sampling interval of $0.152\text{ m}$ after excluding unlogged casing-shoe gaps) are allocated to model training and offset-well prior compilation. Two geographically separated wells are held out as independent blind test wells: Well 1 (15-9-15, comprising 13,290 valid depth samples spanning $1149.6\text{--}3198.8\text{ m}$, Fig.~\ref{fig:dataset_overview}) and Well 2 (15-9-17, comprising 13,058 valid depth samples spanning $1012.6\text{--}3110.5\text{ m}$), yielding 26,348 blind evaluation samples and 87,709 total depth samples. Six conventional wireline curves (GR, CALI, SP, NPHI, RMED, and RDEP) serve as auxiliary inputs $\mathbf{X}$, while bulk density (DEN, $\text{g/cm}^3$, corresponding to FORCE 2020 \texttt{RHOB}) and compressional acoustic transit time (AC, converted to SI units $\mu\text{s/m} = \texttt{DTC} / 0.3048$ from raw $\mu\text{s/ft}$) serve as the reconstruction targets $\mathbf{Y}$ across six consolidated petrophysical regimes $c_t \in \{0, \dots, 5\}$ (Marine Shale, Clean Reservoir Sandstone, Heterolithic Sandstone--Shale, Carbonate and Chalk, Argillaceous Marl, and Volcanic Tuff and Evaporite).

Because single-sample random dropouts ($\sim 0.15\text{ m}$) can be trivially recovered by linear interpolation and do not reflect real wireline acquisition failures \cite{darling2005well,feng2021imputation,zhang2024well,gama2025imputation}, we follow established petrophysical imputation literature \cite{pham2020missing,zhang2024well,al2025well,gama2025imputation} to evaluate two primary field failure protocols on blind test Well 1 and Well 2, supplemented by a continuous $20\%\text{--}100\%$ missingness spectrum in Section~\ref{sec:sensitivity}: (1) the \textbf{M50 Protocol (50\% Contiguous Reservoir-Block Missingness)}, which masks a continuous 50\% central multi-formation block ($6,645$ samples spanning $1654.6\text{--}2690.0\text{ m}$ in Well 1 and $6,529$ samples spanning $1608.6\text{--}2600.9\text{ m}$ in Well 2) while preserving the upper and lower 25\% observed shoulders to model intermediate-run borehole washouts and tool sticking; and (2) the \textbf{M100 Protocol (100\% Full-Well Blind Reconstruction)}, which masks 100\% of DEN and AC across the entire blind wellbore ($>2000\text{ m}$ continuous absence) to model vintage or cost-constrained wells where nuclear density or sonic tools were omitted entirely. \textbf{GeoLog-Agent} is benchmarked against nine mainstream deep learning baselines (\textbf{DLinear} \cite{zeng2023dlinear}, \textbf{LSTM} and \textbf{BiLSTM} \cite{zhang2024well}, \textbf{Transformer} \cite{ashish2017attention,lin2023missing}, \textbf{Informer} \cite{zhou2021informer}, \textbf{Autoformer} \cite{wu2021autoformer}, \textbf{PatchTST} \cite{nie2023patchtst}, \textbf{TimesNet} \cite{wu2023timesnet}, and unconditional \textbf{GAN} \cite{al2025well}), all trained using the AdamW optimizer ($\mathrm{lr} = 5\times 10^{-4}$, sliding window $W = 256$). Across three random seeds ($\{42, 1024, 2025\}$), \textbf{GeoLog-Agent} maintains high stability ($\mathrm{SD}(R^2) \le 0.0024$; main tables report the canonical seed-42 trajectories), and paired Wilcoxon signed-rank tests over non-overlapping $20\text{ m}$ stratigraphic blocks ($N_{\mathrm{block}} = 102$ under M50 and $204$ under M100) confirm $p < 0.0001$ across all comparisons.

\subsection{Comparative Experiments}

Table~\ref{tab:comp_summary} summarizes the overall quantitative comparative performance of all ten models across blind test Well 1 (15-9-15) and Well 2 (15-9-17) under both the M50 and M100 protocols, while complete per-well and per-curve MAE, RMSE, and $R^2$ breakdowns are provided in Appendix Table~\ref{tab:per_well_breakdown}. Across all eight blind-well evaluation scenarios, \textbf{GeoLog-Agent} ranks first on every evaluation metric, achieving an overall eight-scenario mean $R^2$ of $0.9307$ (and a peak $R^2$ of $0.9713$), which exceeds the nine-baseline average ($0.8387$) by $0.0920$ in $R^2$ (a $30.97\%$ average RMSE reduction) and surpasses the strongest baseline (unconditional GAN at $0.9033$ and Informer at $0.8809$) by $0.0274$ to $0.0498$ in $R^2$ (a $16.25\%$ to $24.05\%$ average RMSE reduction).

\begin{table*}[!t]
\centering
\caption{Quantitative comparison of well-log reconstruction across blind test Well 1 (15-9-15) and Well 2 (15-9-17) under M50 and M100 protocols (two-well mean $R^2$, DEN RMSE in $\text{g/cm}^3$, and AC RMSE in $\mu\text{s/m}$; per-well breakdowns are in Table~\ref{tab:per_well_breakdown}). Bold indicates the best result.}
\label{tab:comp_summary}
\footnotesize
\setlength{\tabcolsep}{4.0pt}
\renewcommand{\arraystretch}{1.15}
\resizebox{\textwidth}{!}{%
\begin{tabular}{lccccccc}
\toprule
\multirow{2}{*}{\textbf{Model}} & \multicolumn{3}{c}{\textbf{M50 Protocol (50\% Contiguous Missing)}} & \multicolumn{3}{c}{\textbf{M100 Protocol (100\% Full-Well Blind)}} & \multirow{2}{*}{\textbf{Overall Mean $\mathbf{R^2} \uparrow$}} \\
\cmidrule(lr){2-4} \cmidrule(lr){5-7}
 & \textbf{Mean $\mathbf{R^2} \uparrow$} & \textbf{DEN RMSE $\downarrow$} & \textbf{AC RMSE $\downarrow$} & \textbf{Mean $\mathbf{R^2} \uparrow$} & \textbf{DEN RMSE $\downarrow$} & \textbf{AC RMSE $\downarrow$} & \\
\midrule
DLinear     & 0.7893 & 0.1254 & 38.7793 & 0.5573 & 0.1754 & 59.7646 & 0.6733 \\
Autoformer  & 0.8588 & 0.0773 & 41.7653 & 0.6347 & 0.1320 & 64.5900 & 0.7468 \\
LSTM        & 0.8955 & 0.0750 & 34.5233 & 0.8054 & 0.1022 & 47.2646 & 0.8505 \\
BiLSTM      & 0.9057 & 0.0715 & 32.8235 & 0.8220 & 0.0955 & 46.2017 & 0.8638 \\
TimesNet    & 0.9166 & 0.0683 & 30.3651 & 0.8249 & 0.0970 & 44.9000 & 0.8708 \\
PatchTST    & 0.9178 & 0.0664 & 30.8215 & 0.8394 & 0.0900 & 44.1339 & 0.8786 \\
Transformer & 0.9192 & 0.0667 & 30.1769 & 0.8422 & 0.0907 & 43.0255 & 0.8807 \\
Informer    & 0.9209 & 0.0663 & 29.7026 & 0.8409 & 0.0910 & 43.2882 & 0.8809 \\
GAN         & 0.9357 & 0.0602 & 26.6093 & 0.8710 & 0.0819 & 38.9738 & 0.9033 \\
\textbf{GeoLog-Agent} & \textbf{0.9552} & \textbf{0.0490} & \textbf{22.6147} & \textbf{0.9061} & \textbf{0.0686} & \textbf{33.6994} & \textbf{0.9307} \\
\bottomrule
\end{tabular}%
}
\end{table*}

Under the M50 protocol (Table~\ref{tab:comp_summary} and Appendix Table~\ref{tab:per_well_breakdown}), \textbf{GeoLog-Agent} attains a mean $R^2$ of $0.9552$ (with DEN $R^2 = 0.9484$ on Well 1 and $0.9646$ on Well 2, and AC $R^2 = 0.9366$ on Well 1 and $0.9713$ on Well 2; when evaluated strictly against the masked-interval sample variance $M_{t,j}=1$, \textbf{GeoLog-Agent} likewise ranks first on all four masked-only evaluations with $R^2_{\mathrm{mask}} = 0.9079$, $0.8737$, $0.8751$, and $0.9317$). Relative to the strongest sequence baselines (PatchTST, Informer, and Transformer), \textbf{GeoLog-Agent} reduces M50 DEN RMSE by $18.5\%$ on Well 1 and $32.8\%$ on Well 2 ($14.4\%$ and $23.1\%$ relative to GAN) and reduces M50 AC RMSE by $21.2\%$ on Well 1 and $28.4\%$ on Well 2 ($12.5\%$ and $19.4\%$ relative to GAN). Under the 100\% full-well blind reconstruction protocol (M100, Table~\ref{tab:comp_summary} and Appendix Table~\ref{tab:per_well_breakdown}), where no in-well target observations exist to anchor compaction trends, the performance advantage of \textbf{GeoLog-Agent} widens further: it achieves a mean M100 $R^2$ of $0.9061$ (DEN $R^2 = 0.9235$ and $0.9068$; AC $R^2 = 0.8511$ and $0.9430$), improving $R^2$ by $0.0412$ to $0.0846$ over the strongest per-scenario sequence baselines ($0.0412$ to $0.0886$ across PatchTST, Informer, and Transformer, reducing RMSE by $20.1\%$ to $24.2\%$) and by $0.0225$ to $0.0463$ over unconditional GAN (reducing RMSE by $12.7\%$ to $18.3\%$).

\begin{figure*}[!htbp]
\centering
\subfloat[M50 AC profiles in Well 1 (15-9-15)]{\includegraphics[width=0.475\textwidth]{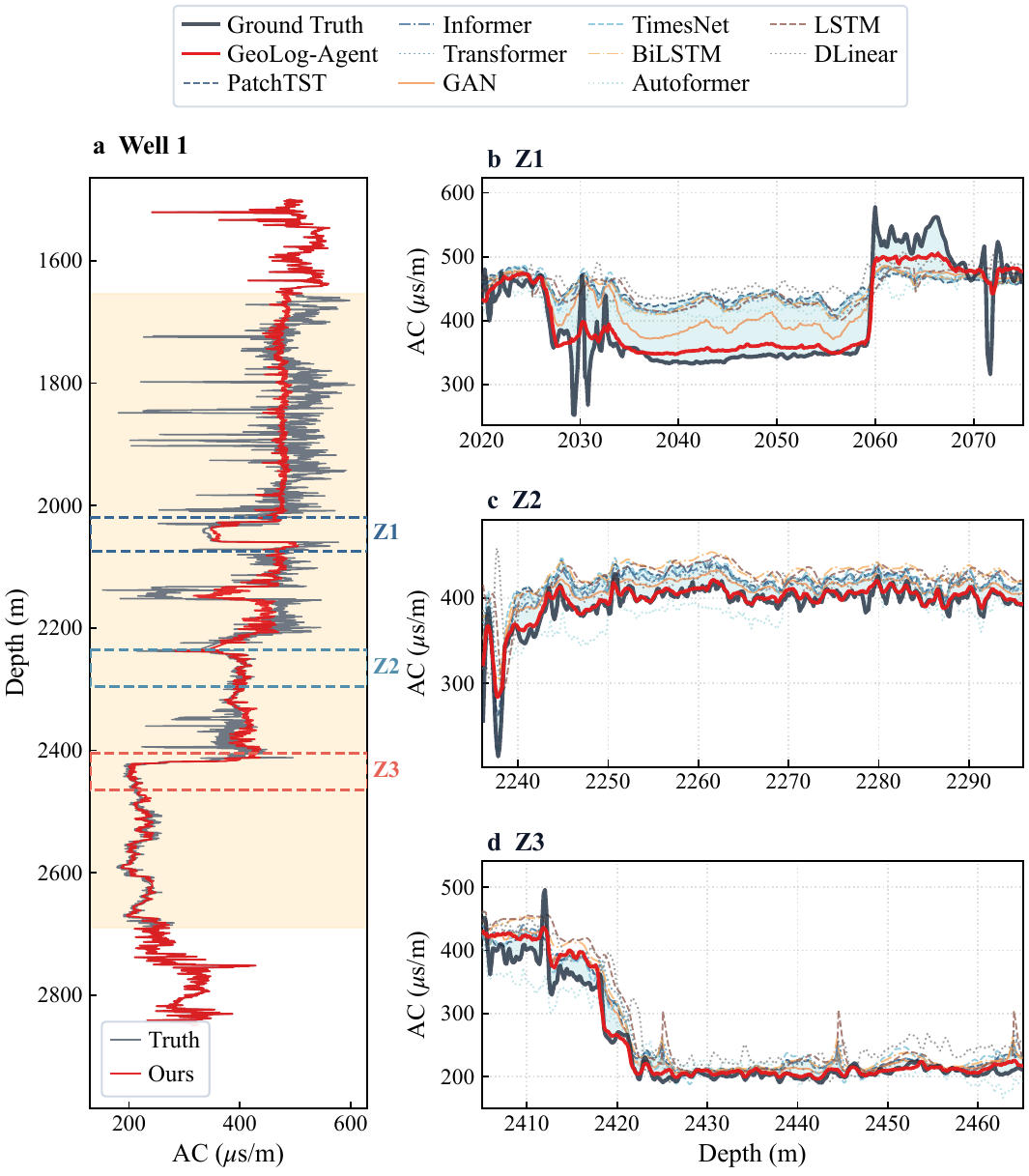}%
\label{fig:zoom_m50_ac}}
\hfil
\subfloat[M50 DEN profiles in Well 1 (15-9-15)]{\includegraphics[width=0.475\textwidth]{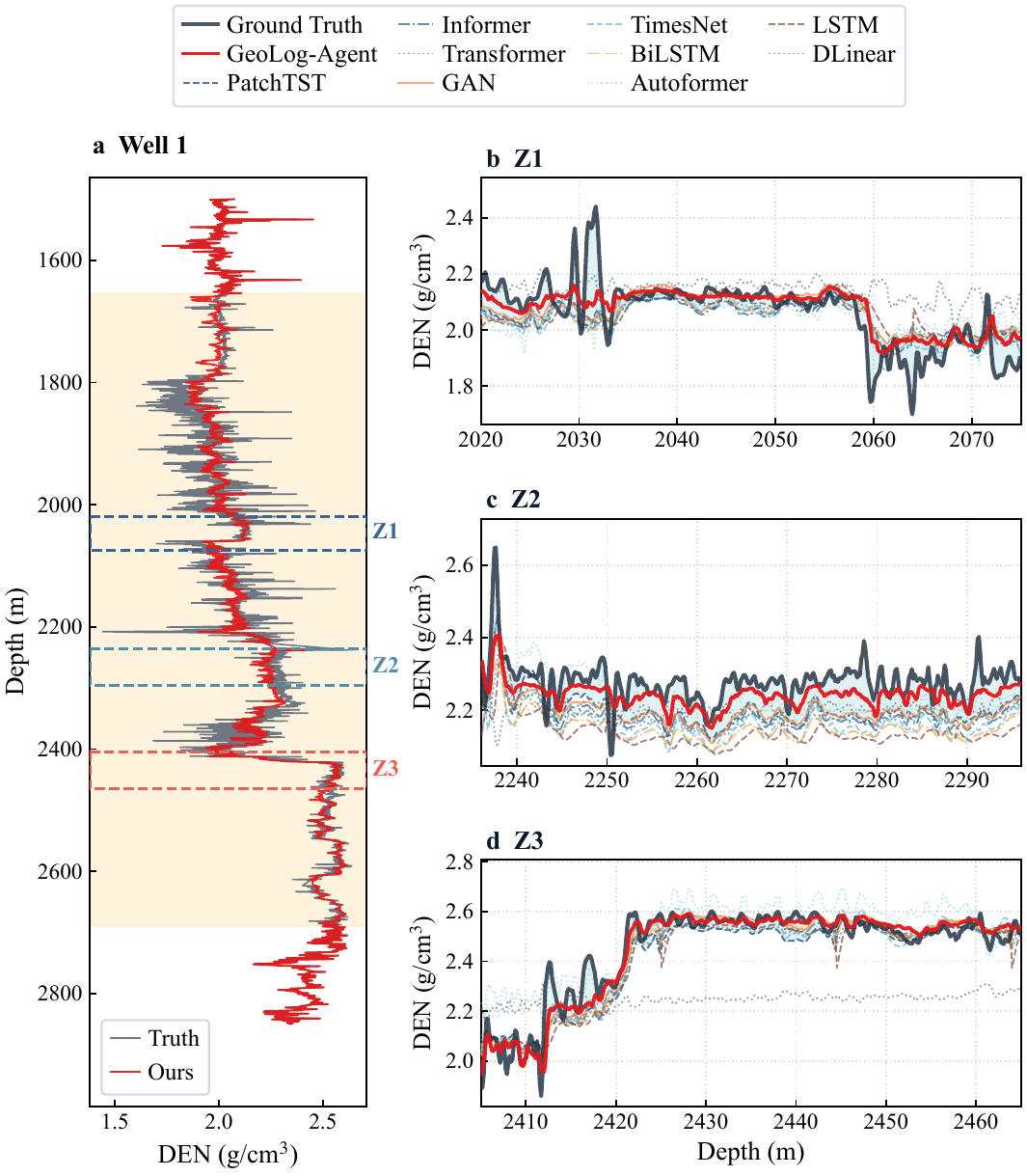}%
\label{fig:zoom_m50_den}}
\\[-2pt]
\subfloat[M100 AC profiles in Well 2 (15-9-17)]{\includegraphics[width=0.475\textwidth]{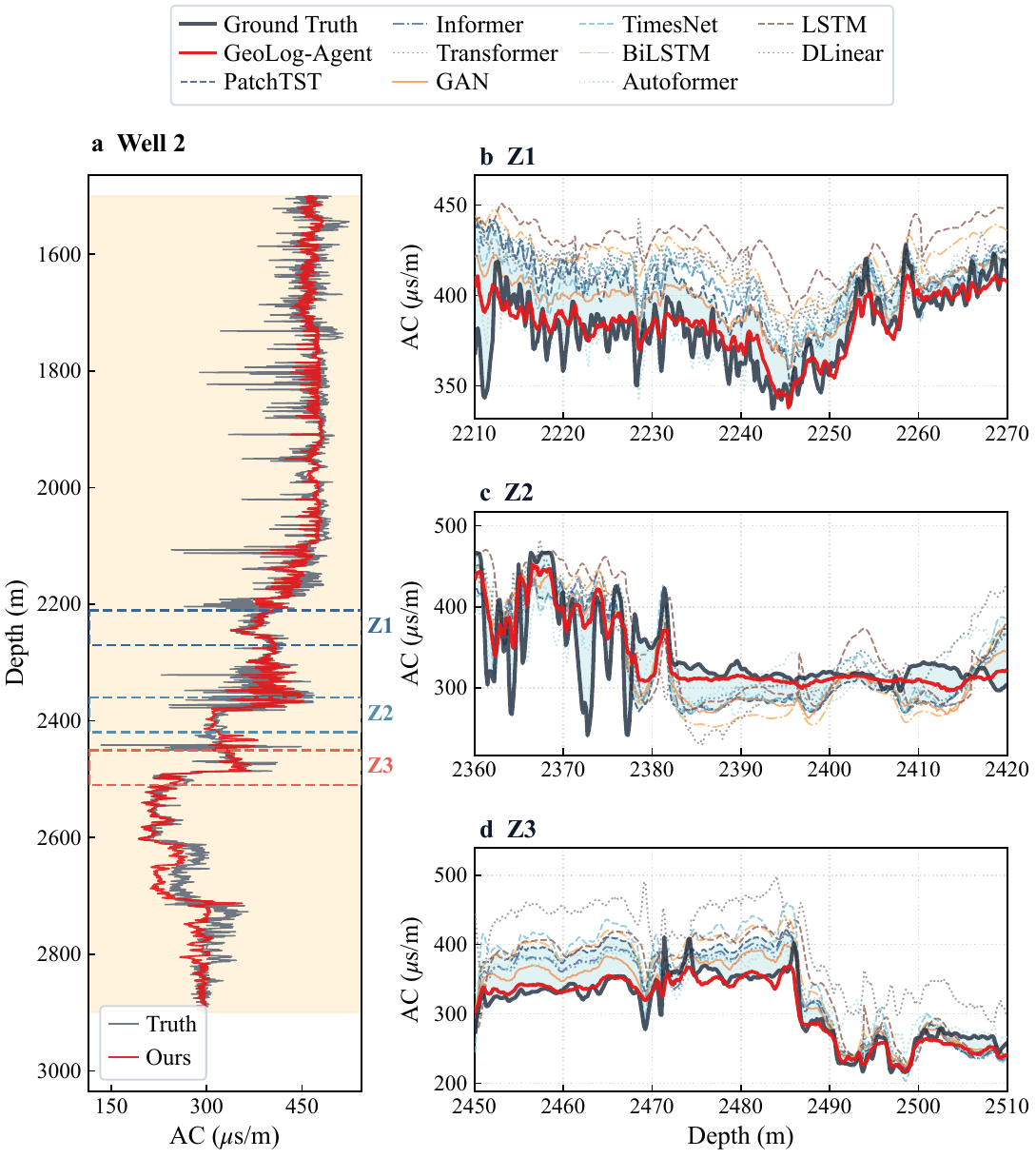}%
\label{fig:zoom_m100_ac}}
\hfil
\subfloat[M100 DEN profiles in Well 2 (15-9-17)]{\includegraphics[width=0.475\textwidth]{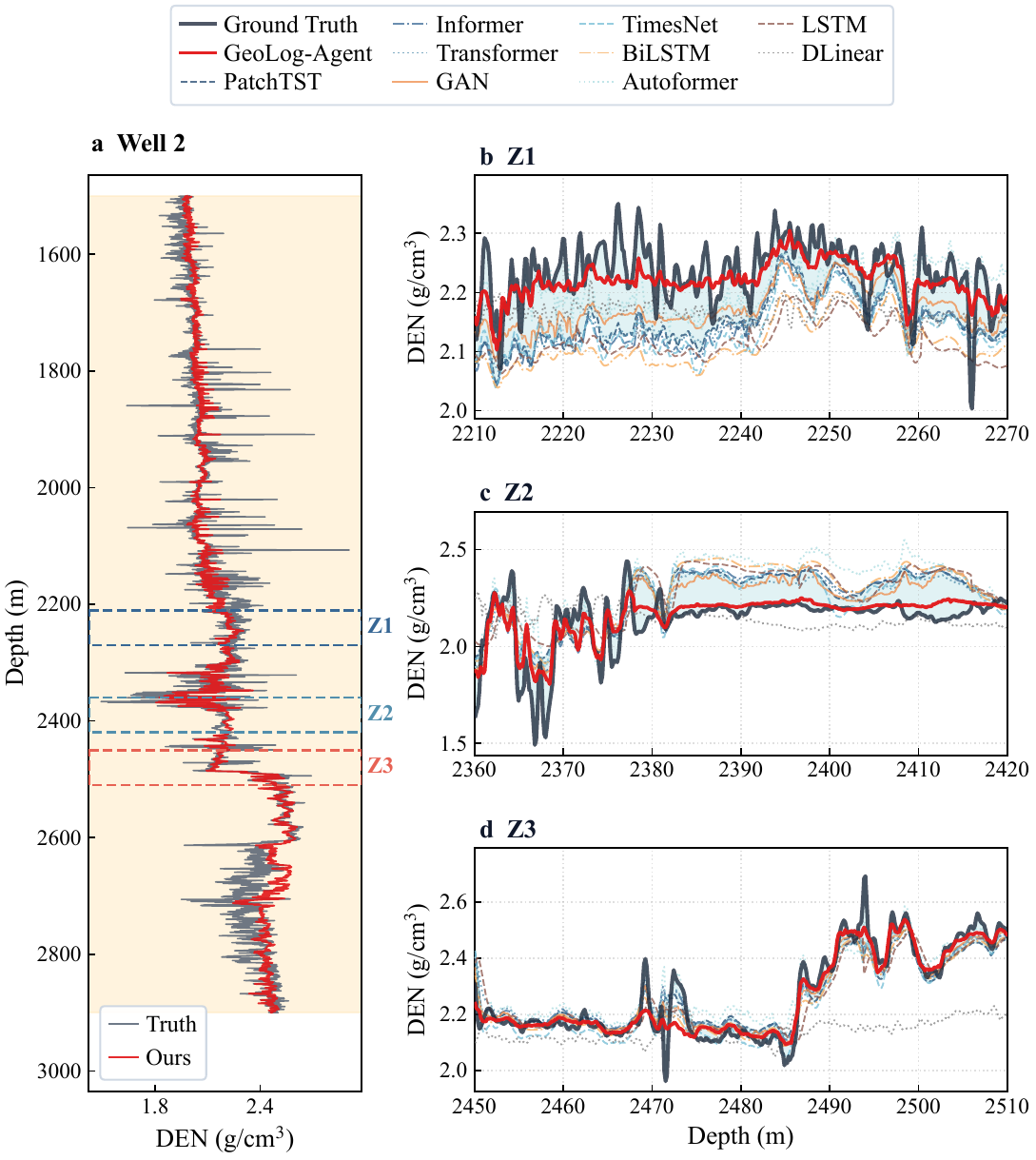}%
\label{fig:zoom_m100_den}}
\caption{Multi-window log reconstruction across all ten models: (\textbf{a}, \textbf{b}) Well 1 under M50 across windows Z1 ($2020\text{--}2075\text{ m}$), Z2 ($2236\text{--}2296\text{ m}$), and Z3 ($2405\text{--}2465\text{ m}$); (\textbf{c}, \textbf{d}) Well 2 under M100 across windows Z1 ($2210\text{--}2270\text{ m}$), Z2 ($2360\text{--}2420\text{ m}$), and Z3 ($2450\text{--}2510\text{ m}$).}
\label{fig:zoom_all_protocols}
\end{figure*}

\begin{figure*}[!t]
\centering
\subfloat[Predicted-versus-true correlation scatter]{\includegraphics[width=0.485\textwidth]{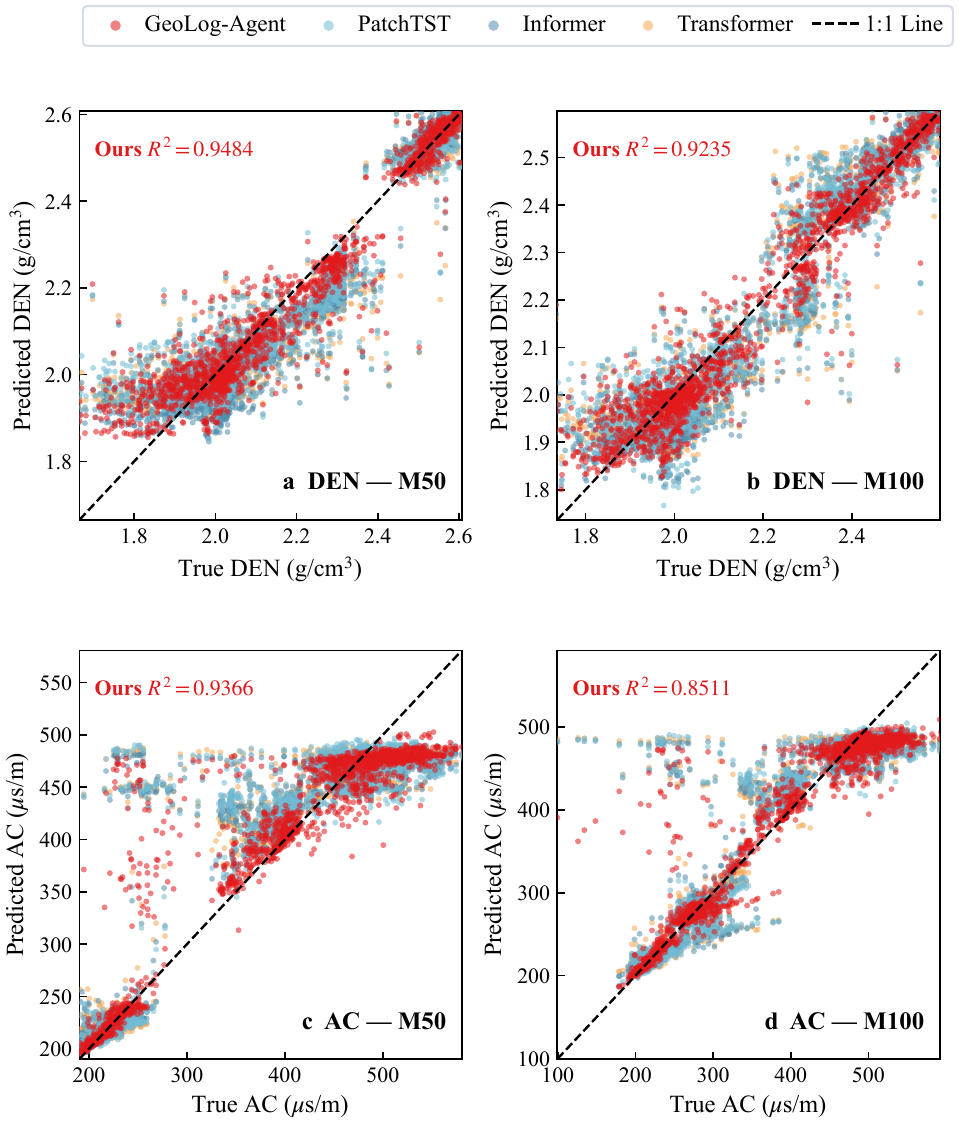}%
\label{fig:scatter_overlay}}
\hfil
\subfloat[Pointwise absolute error distributions]{\includegraphics[width=0.495\textwidth]{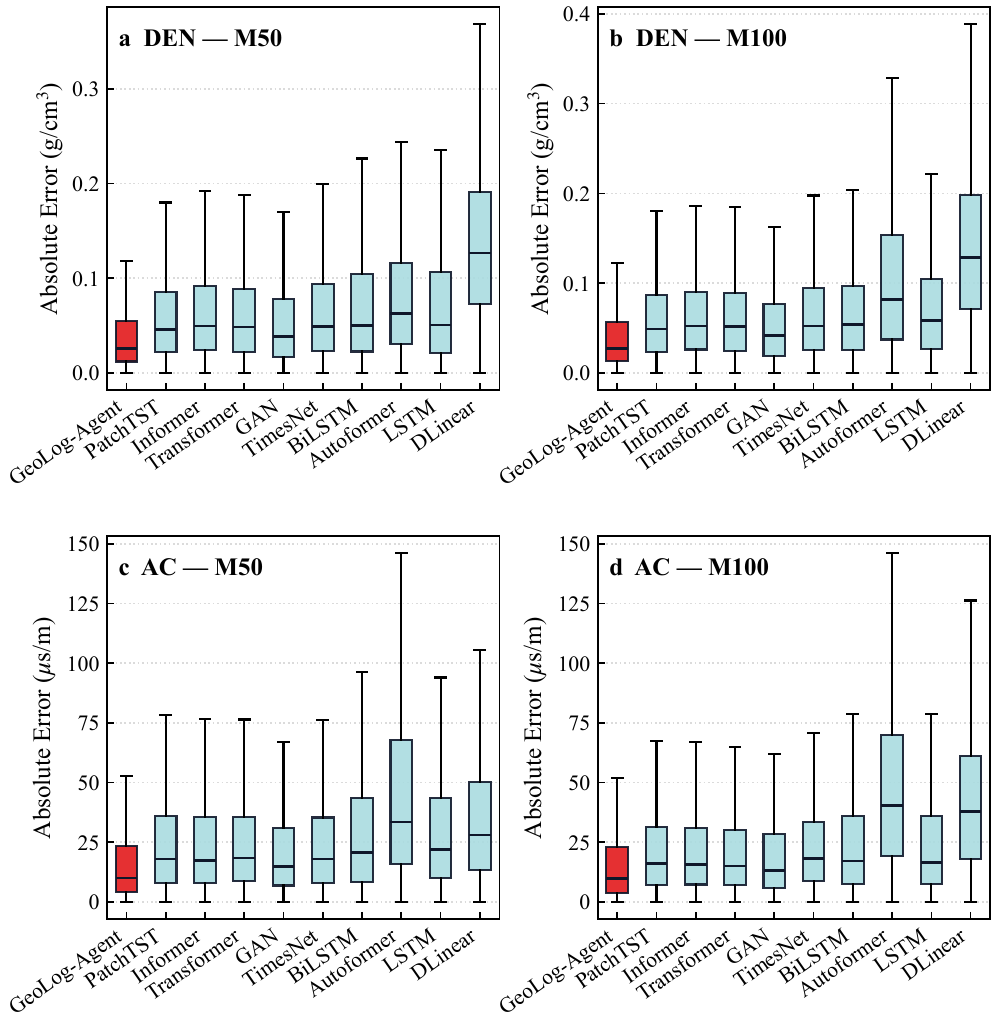}%
\label{fig:error_boxplots}}
\caption{Cross-well correlation fidelity and error distributions under M50 and M100: (\textbf{a}) predicted-versus-true DEN and AC scatters for the top four models, and (\textbf{b}) pointwise absolute error boxplots across all ten models.}
\label{fig:comparison_profiles}
\end{figure*}

Figures~\ref{fig:dataset_overview}b, \ref{fig:zoom_all_protocols}, and \ref{fig:comparison_profiles} illustrate the geological basis of these improvements across blind test Well 1 and Well 2 (supplementary grouped bar charts, hexbin scatter matrices, and six-axis radar profiles are provided in Appendix Figs.~\ref{fig:bar_comparisons}--\ref{fig:radar_appendix}). In the selective zoom across the Rogaland Group ($2236\text{--}2326\text{ m}$ in Well 1 and $2210\text{--}2300\text{ m}$ in Well 2, Fig.~\ref{fig:dataset_overview}b and Fig.~\ref{fig:zoom_all_protocols}), cemented volcanic tuff beds and heterolithic transitions in the Balder and Sele Formations produce sharp density and sonic excursions. Purely numerical sequence models (PatchTST, Informer, and Transformer) exhibit systematic baseline drift across these intervals (shaded blue error bands), resulting in local DEN RMSE of $0.088\text{--}0.109\text{ g/cm}^3$ and AC RMSE of $23.9\text{--}25.9\text{ }\mu\text{s/m}$ in Well 1 ($0.091\text{--}0.093\text{ g/cm}^3$ and $24.0\text{--}29.5\text{ }\mu\text{s/m}$ in Well 2). In contrast, \textbf{GeoLog-Agent} tracks the true petrophysical trajectory closely, achieving a local DEN RMSE of $0.051\text{ g/cm}^3$ ($1.7\times\text{--}2.1\times$ error reduction) and AC RMSE of $8.7\text{ }\mu\text{s/m}$ ($2.7\times\text{--}3.0\times$ error reduction) in Well 1, and $0.037\text{ g/cm}^3$ ($\approx 2.5\times$ reduction) and $10.0\text{ }\mu\text{s/m}$ ($2.4\times\text{--}3.0\times$ reduction) in Well 2 under M100. Similarly, in window Z1 of Well 1 ($2020\text{--}2075\text{ m}$, Fig.~\ref{fig:zoom_all_protocols}a,b), where the wellbore penetrates the clean turbiditic sandstone reservoir of the Frigg Formation ($2028\text{--}2060\text{ m}$, $\mathrm{AC} \approx 340\text{ }\mu\text{s/m}$), conventional sequence baselines remain trapped near the overlying Utsira mudstone baseline ($\mathrm{AC} \approx 430\text{ }\mu\text{s/m}$, PatchTST RMSE $= 78.1\text{ }\mu\text{s/m}$), whereas \textbf{GeoLog-Agent} accurately reconstructs the entire $32\text{ m}$ sandstone reservoir boxcar (RMSE $= 28.0\text{ }\mu\text{s/m}$, a $2.8\times$ error reduction).

\subsection{Ablation Experiments}

\begin{figure*}[!t]
\centering
\subfloat[Cross-scenario $R^2$ score heatmap across all ten models]{\includegraphics[width=0.95\textwidth]{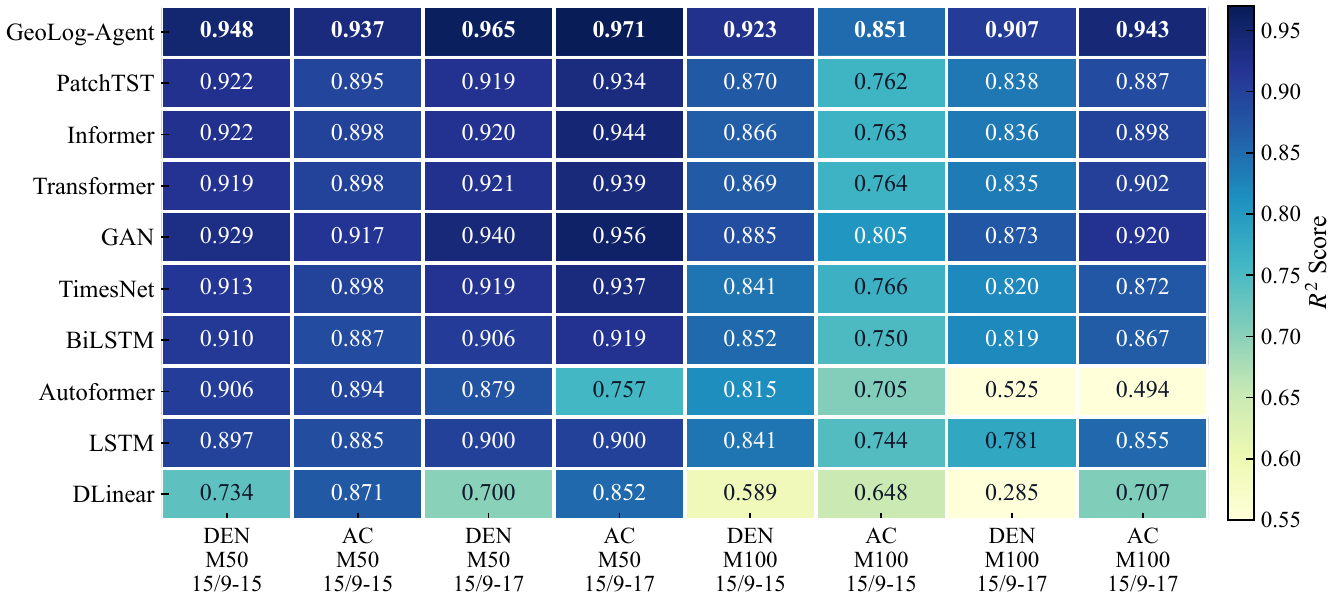}%
\label{fig:heatmap_r2}}\\[4pt]
\subfloat[\textbf{Geo-Harness} plug-and-play gains and ablation hierarchy]{\includegraphics[width=0.95\textwidth]{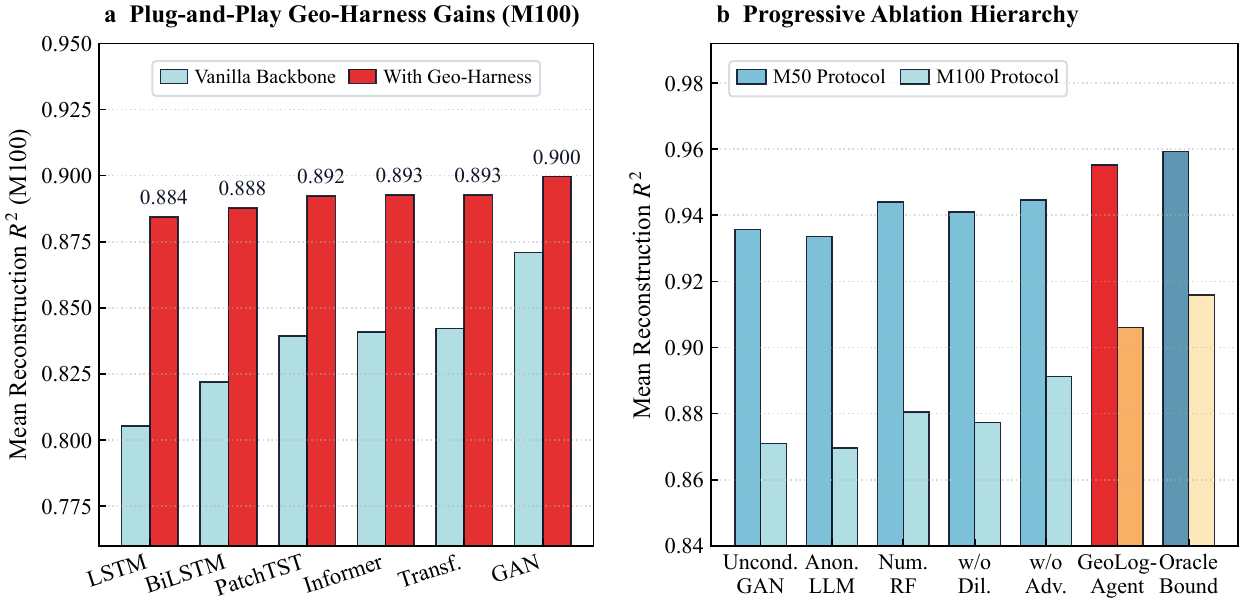}%
\label{fig:plug_and_play_gains}}
\caption{Benchmark $R^2$ matrix, cross-backbone transferability, and component ablation: (\textbf{a}) $R^2$ heatmap across all eight scenarios, and (\textbf{b}) M100 $R^2$ gains when attaching \textbf{Geo-Harness} to six conventional backbones (Panel a) alongside the seven-tier ablation hierarchy of \textbf{GeoLog-Agent} (Panel b).}
\label{fig:ablation_and_harness}
\end{figure*}

To evaluate the individual contribution of each methodological component and verify the role of pre-trained geological knowledge in \textbf{GeoLog-Agent}, Table~\ref{tab:ablation} and Fig.~\ref{fig:ablation_and_harness} examine seven progressive configurations across both blind test wells under M50 and M100: (1) Unconditional GAN ($\mathbf{E}=\mathbf{0}$), (2) \textbf{GeoLog-Agent} with Anonymized LLM Prior (regional formation metadata stripped from the Stage~I prompt), (3) \textbf{GeoLog-Agent} with Numerical RF Prior (supervised Random Forest trained on the six conventional logs), (4) \textbf{GeoLog-Agent} without Multi-Scale Dilation ($r_l=1$), (5) \textbf{GeoLog-Agent} without Adversarial Loss ($L_1$ regression only), (6) the full proposed \textbf{GeoLog-Agent}, and (7) the theoretical Oracle True-Label Ceiling ($\mathbf{c}_{\mathrm{true}}$).

\begin{table*}[!t]
\centering
\caption{Component-wise ablation study of Stage~I geological prior reasoning and Stage~II architectural modules in \textbf{GeoLog-Agent} across blind test Well 1 (15-9-15) and Well 2 (15-9-17) under M50 and M100 protocols. Bold values denote the best non-oracle performance; italicized rows indicate the theoretical Oracle upper bound.}
\label{tab:ablation}
\footnotesize
\setlength{\tabcolsep}{3.5pt}
\renewcommand{\arraystretch}{1.15}
\resizebox{\textwidth}{!}{%
\begin{tabular}{llcccccc}
\toprule
\multirow{2}{*}{\textbf{Scenario}} & \multirow{2}{*}{\textbf{Ablation Configuration}} & \multicolumn{3}{c}{\textbf{Well 1 (15-9-15)}} & \multicolumn{3}{c}{\textbf{Well 2 (15-9-17)}} \\
\cmidrule(lr){3-5} \cmidrule(lr){6-8}
 & & \textbf{DEN} $\mathbf{R^2} \uparrow$ & \textbf{AC} $\mathbf{R^2} \uparrow$ & \textbf{AC MAE} $\downarrow$ & \textbf{DEN} $\mathbf{R^2} \uparrow$ & \textbf{AC} $\mathbf{R^2} \uparrow$ & \textbf{AC MAE} $\downarrow$ \\
\midrule
\multirow{7}{*}{\textbf{M50}}
 & 1. Unconditional GAN (w/o Geo-Harness Prior)       & 0.9295 & 0.9171 & 15.2138 & 0.9403 & 0.9558 &  9.1682 \\
 & 2. GeoLog-Agent (Anonymized LLM Prior)             & 0.9372 & 0.9161 & 15.5723 & 0.9316 & 0.9494 &  9.5689 \\
 & 3. GeoLog-Agent (Numerical RF Prior)               & 0.9385 & 0.9228 & 14.3854 & 0.9493 & 0.9652 &  8.0631 \\
 & 4. GeoLog-Agent (w/o Multi-Scale Dilation, $r_l=1$)& 0.9268 & 0.9236 & 14.1076 & 0.9520 & 0.9616 &  8.1582 \\
 & 5. GeoLog-Agent (w/o Adversarial Loss, $L_1$ Only) & 0.9412 & 0.9298 & 13.6587 & 0.9482 & 0.9590 &  8.2259 \\
 & \textbf{6. GeoLog-Agent (Full Proposed Agent)}     & \textbf{0.9484} & \textbf{0.9366} & \textbf{12.5583} & \textbf{0.9646} & \textbf{0.9713} & \textbf{6.8396} \\
 \cmidrule(lr){2-8}
 & \textit{7. GeoLog-Agent (Oracle True-Label Ceiling)} & \textit{0.9509} & \textit{0.9396} & \textit{12.2643} & \textit{0.9692} & \textit{0.9773} & \textit{6.3804} \\
\midrule
\multirow{7}{*}{\textbf{M100}}
 & 1. Unconditional GAN (w/o Geo-Harness Prior)       & 0.8854 & 0.8048 & 28.5244 & 0.8731 & 0.9205 & 17.3793 \\
 & 2. GeoLog-Agent (Anonymized LLM Prior)             & 0.8948 & 0.8092 & 29.8496 & 0.8650 & 0.9096 & 18.0247 \\
 & 3. GeoLog-Agent (Numerical RF Prior)               & 0.8968 & 0.8240 & 27.0692 & 0.8785 & 0.9230 & 17.3675 \\
 & 4. GeoLog-Agent (w/o Multi-Scale Dilation, $r_l=1$)& 0.8937 & 0.8118 & 26.7572 & 0.8736 & 0.9303 & 15.5341 \\
 & 5. GeoLog-Agent (w/o Adversarial Loss, $L_1$ Only) & 0.9120 & 0.8392 & 25.8108 & 0.8852 & 0.9286 & 15.5465 \\
 & \textbf{6. GeoLog-Agent (Full Proposed Agent)}     & \textbf{0.9235} & \textbf{0.8511} & \textbf{24.2737} & \textbf{0.9068} & \textbf{0.9430} & \textbf{13.7934} \\
 \cmidrule(lr){2-8}
 & \textit{7. GeoLog-Agent (Oracle True-Label Ceiling)} & \textit{0.9278} & \textit{0.8561} & \textit{23.9340} & \textit{0.9243} & \textit{0.9557} & \textit{12.6202} \\
\bottomrule
\end{tabular}%
}
\end{table*}

As shown in Table~\ref{tab:ablation} and Fig.~\ref{fig:ablation_and_harness}c, every component within \textbf{GeoLog-Agent} contributes substantially to final reconstruction accuracy. Removing Stage~I geological priors altogether (Row 1, Unconditional GAN) reduces mean $R^2$ from $0.9552$ to $0.9357$ under M50 and from $0.9061$ to $0.8710$ under M100. More importantly, when the LLM is queried with anonymized numerical summaries stripped of regional formation names (Row 2, Anonymized LLM Prior), mean $R^2$ drops to $0.9336$ under M50 and $0.8697$ under M100, confirming that the gain stems from pre-trained stratigraphic knowledge rather than generic LLM prompting. Similarly, replacing the pre-trained LLM reasoner with a supervised Random Forest classifier trained on the six conventional logs (Row 3) lowers M100 mean $R^2$ to $0.8806$ because purely numerical classification collapses on polarity-reversing volcanic tuff layers. In Stage~II, removing multi-scale cyclic dilation (Row 4) or conditional adversarial training (Row 5) degrades high-frequency waveform sharpness across formation boundaries, reducing M100 mean $R^2$ to $0.8774$ and $0.8913$, respectively. Overall, the full \textbf{GeoLog-Agent} (Row 6) recovers $78.1\%$ to $82.9\%$ of the mean theoretical Oracle $R^2$ gain over unconditional GAN across both wells (and up to $88.3\%$ to $90.3\%$ on Well 1, Row 7) without requiring target-well core labels.

\subsection{Sensitivity Experiments}
\label{sec:sensitivity}

\begin{figure*}[!t]
\centering
\includegraphics[width=0.98\textwidth]{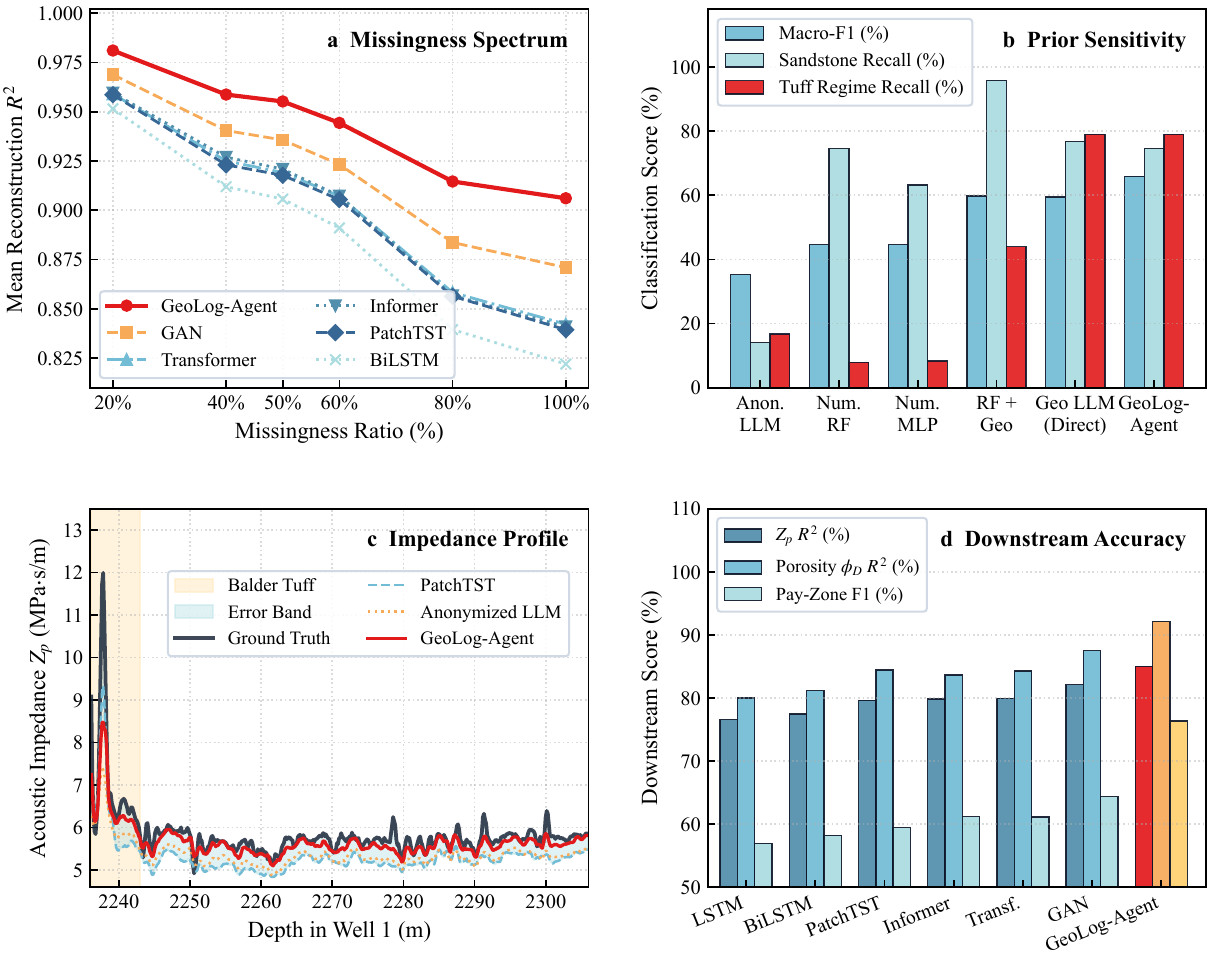}
\caption{Sensitivity analysis and downstream reservoir evaluation: (\textbf{a}) $R^2$ across the $20\%\text{--}100\%$ missingness spectrum, (\textbf{b}) Stage~I prior sensitivity on Macro-F1 and sandstone/tuff recall, (\textbf{c}) Balder--Sele acoustic impedance ($Z_p$) profile in Well 1 ($2236\text{--}2306\text{ m}$), and (\textbf{d}) downstream $Z_p$, porosity ($\phi_D$), and pay-zone Macro-F1 accuracy.}
\label{fig:sensitivity_and_downstream}
\end{figure*}

To evaluate the universality, robustness, and practical engineering value of the proposed paradigm, we conduct four complementary sensitivity and application analyses (Table~\ref{tab:plug_and_play}, Fig.~\ref{fig:ablation_and_harness}b, Fig.~\ref{fig:sensitivity_and_downstream}, and Appendix Table~\ref{tab:downstream_app}):
\begin{enumerate}
  \item \textbf{Cross-Backbone Transferability Across Six Deep Learning Models (Table~\ref{tab:plug_and_play} and Fig.~\ref{fig:ablation_and_harness}b):} Attaching \textbf{Geo-Harness} directly to six representative sequence and generative architectures (\textbf{LSTM}, \textbf{BiLSTM}, \textbf{PatchTST}, \textbf{Informer}, \textbf{Transformer}, and \textbf{GAN}) universally boosts blind-well $R^2$ by $0.0162$ to $0.0789$ overall ($0.0288$ to $0.0789$ under M100) and reduces reconstruction RMSE by $12.23\%$ to $26.28\%$ across all scenarios (e.g., elevating M100 mean $R^2$ from $0.8054$ to $0.8843$ for LSTM with a $23.67\%$ RMSE reduction, from $0.8220$ to $0.8878$ for BiLSTM with a $21.41\%$ RMSE reduction, and from $0.8394\text{--}0.8422$ to $0.8923\text{--}0.8928$ across PatchTST, Informer, and Transformer).
  \item \textbf{Continuous Missingness-Ratio Spectrum Sensitivity ($20\%\text{--}100\%$, Fig.~\ref{fig:sensitivity_and_downstream}a):} Evaluates contiguous missing blocks spanning $20\%, 40\%, 50\%, 60\%, 80\%,$ and $100\%$ of each blind wellbore. While conventional sequence models suffer steep non-linear $R^2$ decay once contiguous missingness exceeds $50\%$ (dropping from $\sim 0.955$ at $20\%$ to $0.822\text{--}0.842$ at $100\%$), \textbf{GeoLog-Agent} maintains a flat, resilient trajectory ($R^2 = 0.9811$ at $20\%$, $0.9552$ at $50\%$, and $0.9061$ at $100\%$).
  \item \textbf{Stage-I Geological Prior Completeness Sensitivity (Fig.~\ref{fig:sensitivity_and_downstream}b):} Compares six Stage~I prior strategies strictly without target DEN/AC logs, including 6-log numerical classifiers (Random Forest and MLP), a same-information non-LLM control supplied with one-hot Formation and Group metadata (\texttt{Numerical RF + Geo Metadata}), and pre-trained LLM reasoning variants. Stripping geological metadata (\texttt{Anonymized LLM}) yields $0.00\%$ volcanic tuff regime recall on Well 1, and even when Numerical RF is explicitly given Formation and Group metadata, it recalls only $33.33\%$ (Well 1) and $54.69\%$ (Well 2) of Regime 5 (Volcanic Tuff/Evaporite) because internal tuff-to-shale sub-beds within the Balder Formation remain ambiguous to shallow tree splits. In contrast, \textbf{GeoLog-Agent} elevates wellbore-wide Regime-5 tuff recall to $83.70\%$ on Well 1 and $74.38\%$ on Well 2 ($100.0\%$ recall strictly within the Balder Formation tuff interval itself), outperforming Pure Numerical RF by $26.48$ and $16.12$ percentage points in Macro-F1 (and surpassing same-information \texttt{RF + Geo Metadata} by $9.90$ and $2.43$ percentage points).
  \item \textbf{Downstream Reservoir Engineering Evaluation (Fig.~\ref{fig:sensitivity_and_downstream}c,d and Appendix Table~\ref{tab:downstream_app}):} Propagating the reconstructed DEN ($\hat{\rho}_b$) and AC ($\widehat{\Delta t}$) curves across the unrecorded intervals ($M_{t,j}=1$) into three log-derived geoenergy tasks demonstrates substantial operational gains: on P-wave acoustic impedance inversion ($Z_p = 10^3 \cdot \rho_b / \Delta t$, $\text{MPa}\cdot\text{s/m}$), \textbf{GeoLog-Agent} attains $R^2 = 0.8501$ ($\mathrm{RMSE} = 1.0349\text{ MPa}\cdot\text{s/m}$), reducing impedance RMSE by $15.9\%$ over Transformer, by $10.1\%$ over GAN, and by $39.0\%$ locally across the Balder--Sele volcaniclastic interval ($0.36\text{ MPa}\cdot\text{s/m}$ vs.\ $0.59\text{ MPa}\cdot\text{s/m}$ for PatchTST over $2236\text{--}2306\text{ m}$ in Fig.~\ref{fig:sensitivity_and_downstream}c); on density-derived effective porosity ($\phi_D$), \textbf{GeoLog-Agent} elevates $R^2$ to $0.9216$ and lowers RMSE to $3.59\text{ p.u.}$ ($27.3\%$ reduction vs.\ PatchTST and $20.0\%$ vs.\ GAN); and on four-zone reservoir pay-zone delineation, \textbf{GeoLog-Agent} boosts Macro-F1 to $76.34\%$ (overall accuracy $91.25\%$), improving upon Informer ($61.15\%$) by $15.19$ percentage points and GAN ($64.36\%$) by $11.98$ percentage points.
\end{enumerate}

\begin{table}[!t]
\centering
\caption{Cross-backbone transferability of the pre-trained geological knowledge coupling mechanism (\textbf{Geo-Harness}) attached to six conventional deep learning models (averaged across DEN and AC on blind test Wells 15-9-15 and 15-9-17).}
\label{tab:plug_and_play}
\footnotesize
\setlength{\tabcolsep}{4.5pt}
\renewcommand{\arraystretch}{1.15}
\resizebox{\textwidth}{!}{%
\begin{tabular}{llcccc}
\toprule
\textbf{Protocol} & \textbf{Backbone Architecture} & \textbf{Vanilla $\mathbf{R^2}$} & \textbf{With Geo-Harness $\mathbf{R^2}$} & \textbf{$\mathbf{\Delta R^2}$ Gain} & \textbf{RMSE Reduction (\%)} \\
\midrule
\multirow{6}{*}{\textbf{M50}}
 & LSTM        & 0.8955 & \textbf{0.9424} & 0.0469 & 26.28\% \\
 & BiLSTM      & 0.9057 & \textbf{0.9453} & 0.0396 & 24.54\% \\
 & PatchTST    & 0.9178 & \textbf{0.9479} & 0.0301 & 21.04\% \\
 & Informer    & 0.9209 & \textbf{0.9490} & 0.0281 & 20.30\% \\
 & Transformer & 0.9192 & \textbf{0.9480} & 0.0288 & 20.43\% \\
 & GAN         & 0.9357 & \textbf{0.9519} & 0.0162 & 14.17\% \\
\midrule
\multirow{6}{*}{\textbf{M100}}
 & LSTM        & 0.8054 & \textbf{0.8843} & 0.0789 & 23.67\% \\
 & BiLSTM      & 0.8220 & \textbf{0.8878} & 0.0658 & 21.41\% \\
 & PatchTST    & 0.8394 & \textbf{0.8923} & 0.0529 & 18.81\% \\
 & Informer    & 0.8409 & \textbf{0.8926} & 0.0517 & 18.40\% \\
 & Transformer & 0.8422 & \textbf{0.8928} & 0.0506 & 17.99\% \\
 & GAN         & 0.8710 & \textbf{0.8997} & 0.0288 & 12.23\% \\
\bottomrule
\end{tabular}%
}
\end{table}

\section{Discussion}
\label{sec:discussion}

\subsection{Complementary Synergy of Pre-trained and Deep Models}

Taking our cross-backbone experiments (Table~\ref{tab:plug_and_play}) as an empirical starting point—where coupling pre-trained geological priors consistently elevates six distinct recurrent, attention, and adversarial architectures—reveals a fundamental principle regarding the cognitive division of labor in geoscientific artificial intelligence. Subsurface sedimentary basins are governed simultaneously by two distinct physical scales: at the macro scale, long-term tectonic subsidence, sea-level eustasy, and volcanic or depositional episodes partition the crust into discrete stratigraphic sequences and diagenetic regimes; at the micro scale, sub-meter variations in pore geometry, mineral matrix stiffness, and fluid saturation generate continuous, high-frequency geophysical waveform fluctuations. Historically, computational geoscience has attempted to model both scales using a single class of neural architecture, creating a persistent methodological bottleneck.

Generalist pre-trained foundation models (such as large language models) absorb vast geoscientific literature, basin stratigraphy, and qualitative rock-physics rules during large-scale pre-training, endowing them with strong macro-scale, symbolic, and contextual reasoning capabilities. However, because foundation models operate over discrete tokenized representations, they are structurally ill-suited for synthesizing continuous physical fields at sub-meter spatial resolution without quantization drift. Conversely, specialist deep domain models trained directly on numerical sensor tensors act as high-precision continuous function approximators capable of capturing localized differential operators and fine-scale waveform textures, yet their inductive bias is confined to local numerical correlations within finite training wells, leaving them epistemically blind to basin-scale geological context. Our findings establish that generalist foundation models and specialist deep generative networks are strictly complementary layers of a unified computational hierarchy: the foundation model governs macro-scale geological semantics and regime boundaries in symbolic space, while the deep generative network executes micro-scale continuous physical field synthesis in tensor space. This cross-scale functional decomposition explains why the resulting performance gains are architecture-agnostic and defines a generalizable paradigm for scientific computing across multi-scale physical systems.

\subsection{Knowledge Regularization in Ill-Posed Inverse Problems}

Stepping beyond specific well-log curves or local lithologies, the systematic failure of unguided deep models across complex stratigraphic transitions uncovers a universal theoretical law governing ill-posed inverse problems in subsurface earth exploration. Across a broad family of geophysical and engineering sensing systems—ranging from borehole logging and seismic inversion to gravimetric, electromagnetic, and remote-sensing earth observation—accessible indirect measurements $\mathbf{X}$ map non-bijectively to target subsurface physical states $\mathbf{Y}$. In heterogeneous geological basins, distinct tectono-sedimentary processes and diagenetic pathways frequently converge onto overlapping low-dimensional auxiliary signatures while obeying divergent elastic, density, or fluid-flow constitutive laws, causing the true conditional distribution $p(\mathbf{Y} \mid \mathbf{X})$ to bifurcate into a multimodal distribution across latent geological regimes $c_t \in \{0, \dots, K-1\}$.

When any purely data-driven model is trained under standard empirical risk minimization without macro-scale domain priors, it is mathematically forced to approximate the marginal conditional expectation $\mathbb{E}[\mathbf{Y} \mid \mathbf{X}] = \sum_k p(c_t = k \mid \mathbf{X})\,\mathbb{E}[\mathbf{Y} \mid \mathbf{X}, c_t = k]$ averaged across all competing physical modes. This phenomenon—which we formalize as the \emph{cross-regime mode-averaging trap}—inevitably collapses sharp geological boundaries, fault-zone transitions, and thin high-contrast reservoirs into physically implausible intermediate trajectories, regardless of network depth or attention capacity. Incorporating macro-scale domain priors acts as an epistemic regularizer that resolves this ill-posedness at its mathematical root: by inferring the governing geological regime $\hat{\mathbf{c}}$ from basin-scale contextual cues (such as sequence-stratigraphic topology and regional depositional rules), it decomposes a globally non-unique, high-entropy inverse mapping into locally unimodal, physically well-posed sub-manifolds ($\mathcal{H}(\mathbf{Y} \mid \mathbf{X}, \hat{\mathbf{c}}) \ll \mathcal{H}(\mathbf{Y} \mid \mathbf{X})$). Crucially, injecting these domain priors directly into intermediate neural representations dynamically re-orients the generative manifold across regime boundaries, transforming black-box curve fitting into law-constrained physical synthesis.

\subsection{Broader Implications for Scientific Computing}

The paradigm validated in this study—using pre-trained foundation models to steer deep generative networks—carries far-reaching implications for petroleum exploration, basin-scale reservoir modeling, and emerging geoenergy systems beyond single-well log reconstruction. First, in deep and ultra-deep hydrocarbon exploration across mature and frontier sedimentary basins worldwide, tens of thousands of legacy appraisal and development wells lack complete elastic or density log suites due to historical cost constraints or harsh downhole conditions. By unlocking these incomplete historical archives with geology-consistent physical trajectories, knowledge-guided generative agents provide reliable well-to-seismic calibration anchors for 3D basin-scale seismic inversion, sequence-stratigraphic correlation, and the delineation of subtle stratigraphic and lithological hydrocarbon traps without the prohibitive cost of re-drilling or re-logging. Moreover, in frontier wildcat exploration where formal formation tops have not yet been established, the same symbolic conditioning interface can ingest unstructured mud-logging cutting descriptions, core photographs, drilling mechanics logs, or regional seismic facies attributes to guide continuous subsurface property generation.

Second, this collaborative paradigm addresses a critical bottleneck in the global energy transition, including geological carbon sequestration ($\text{CO}_2$ storage caprock integrity and plume monitoring), deep geothermal reservoir engineering, and underground hydrogen storage. Unlike mature oilfields, these emerging geoenergy sectors operate in data-scarce geological formations where labeled training wells are far too sparse to train standalone deep learning models from scratch, yet where decades of regional geological surveys and rock-physics literature already exist. Pre-trained foundation models provide a zero-shot vehicle to transfer this accumulated human geoscientific knowledge into deep generative models across data-sparse basins. More broadly, for researchers across earth system science and engineering computing, this study demonstrates a general methodological principle: whenever continuous physical fields must be reconstructed or inverted under structural data absence and multi-regime non-stationarity, coupling macro-scale foundation-model domain reasoning with micro-scale deep generative synthesis offers a principled pathway from purely statistical interpolation to context-aware scientific discovery.

\section{Conclusion}
\label{sec:conclusion}

This study addressed a fundamental challenge in subsurface geophysical exploration and ill-posed inverse modeling: purely numerical deep learning models lack macro-scale geological prior knowledge and consequently fall into cross-regime mode averaging when reconstructing physical fields across heterogeneous sedimentary formations. To resolve this limitation, we established a collaborative scientific paradigm—instantiated through the knowledge-conditioned generative agent (\textbf{GeoLog-Agent}) and the universal coupling harness (\textbf{Geo-Harness})—that integrates macro-scale geological prior reasoning in pre-trained large language models with micro-scale continuous waveform synthesis in deep generative networks.

From a theoretical and methodological perspective, our investigation yields two generalizable insights. First, in non-bijective geophysical inverse problems where distinct geological formations produce overlapping auxiliary measurements, macro-scale stratigraphic and rock-physics priors act as an essential epistemic regularizer that partitions globally multimodal mappings into locally well-posed physical sub-manifolds. This mechanism enables \textbf{GeoLog-Agent} to preserve sharp lithological boundaries and physical extremes across complex volcanic, carbonate, and siliciclastic transitions, achieving high blind-well reconstruction fidelity (mean $R^2 = 0.9307$) while recovering over four-fifths of the theoretical Oracle prior bound without target-well retraining. Second, our cross-backbone experiments demonstrate a universal cognitive division of labor between generalist foundation models and specialist deep domain models: delegating macro-scale symbolic regime inference to pre-trained foundation models while reserving continuous high-frequency signal synthesis for deep neural generators consistently elevates recurrent, attention, and generative architectures across wide missingness spectra.

From an engineering and domain perspective, coupling pre-trained geological knowledge with deep generative synthesis prevents local waveform smoothing errors from cascading into downstream seismic-to-well calibration, petrophysical inversion, and hydrocarbon pay-zone evaluation. Beyond wellbore log reconstruction, this work provides a transferable methodological blueprint for petroleum geology, basin-scale digital subsurface modeling, and emerging geoenergy applications—demonstrating how pre-trained foundation models can transform deep generative AI from context-blind numerical fitting into geology-informed scientific computing.

\appendix
\setcounter{table}{0}
\setcounter{figure}{0}

\section{Supplementary Material}
\label{sec:appendix}

\begin{figure*}[!htbp]
\centering
\subfloat[Grouped MAE and RMSE comparison]{\includegraphics[width=0.485\textwidth]{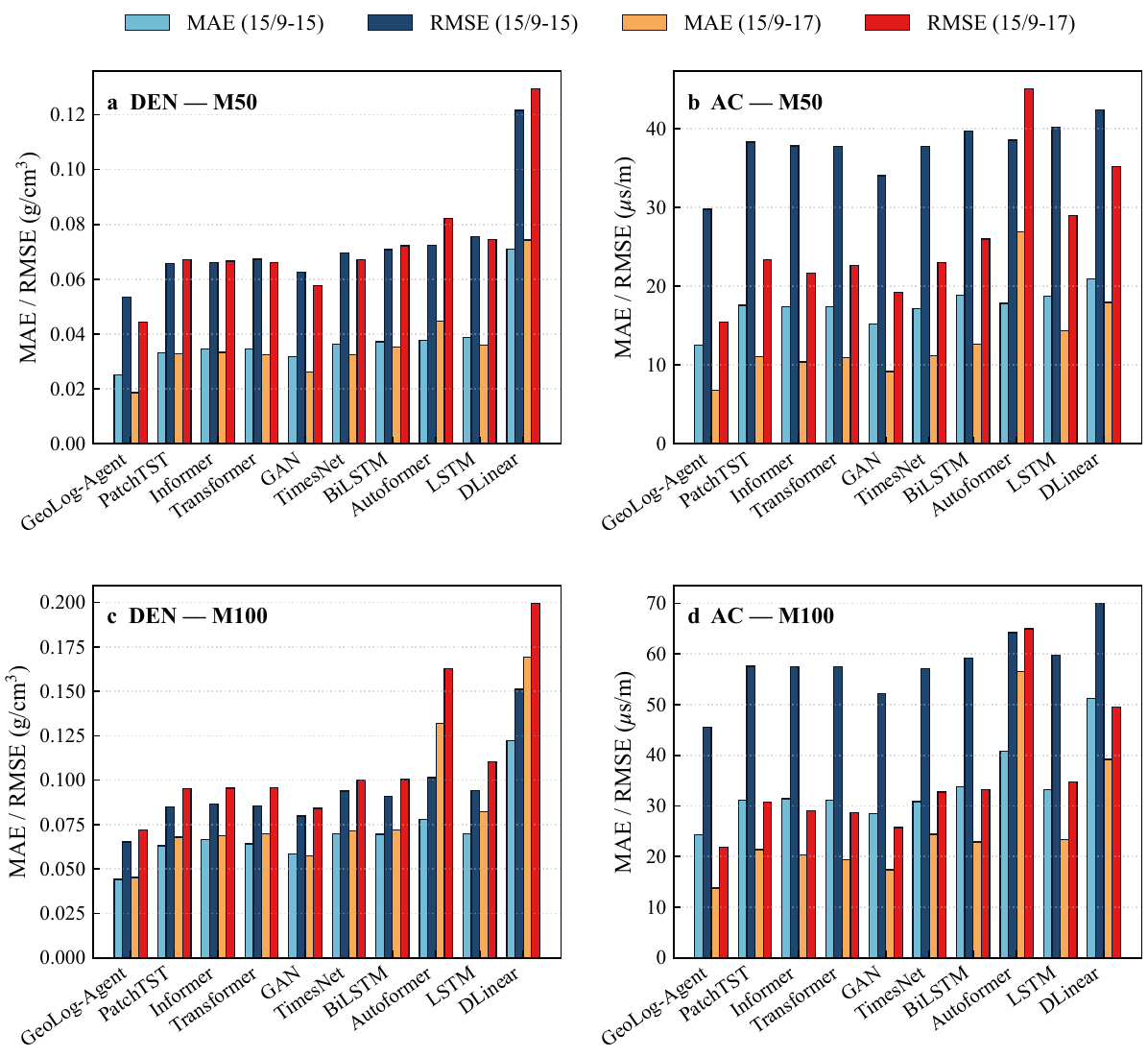}%
\label{fig:bar_mae_rmse}}
\hfil
\subfloat[Grouped bar comparison of $R^2$]{\includegraphics[width=0.485\textwidth]{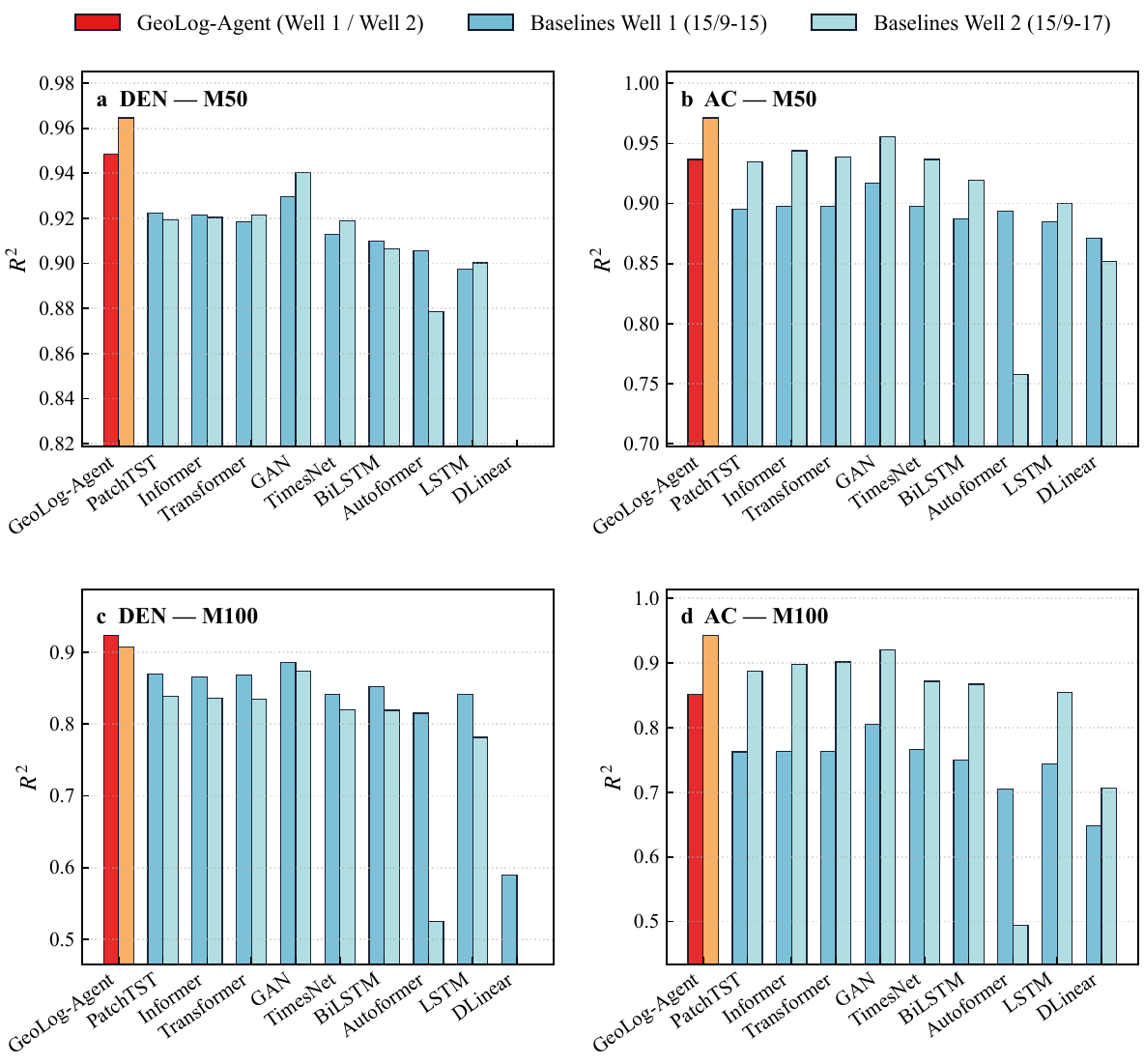}%
\label{fig:bar_r2}}
\caption{Multi-well grouped bar comparisons across M50 and M100 scenarios: (\textbf{a}) MAE and RMSE, and (\textbf{b}) $R^2$ across Well 1 (15-9-15) and Well 2 (15-9-17).}
\label{fig:bar_comparisons}
\end{figure*}

\begin{figure*}[!htbp]
\centering
\subfloat[Well 1 AC hexbin scatter matrix]{\includegraphics[width=0.48\textwidth]{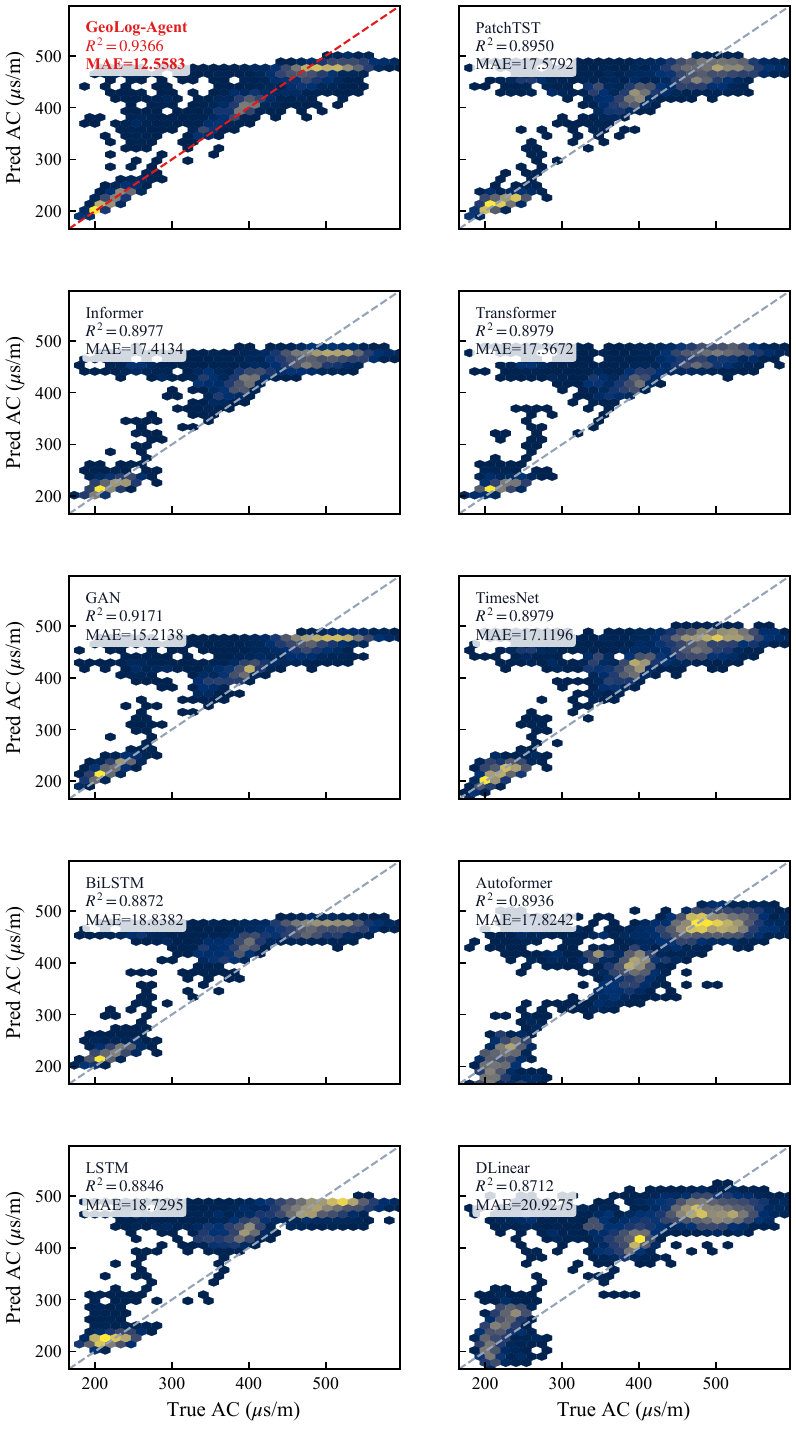}%
\label{fig:scatter_w1_ac}}
\hfil
\subfloat[Well 1 DEN hexbin scatter matrix]{\includegraphics[width=0.48\textwidth]{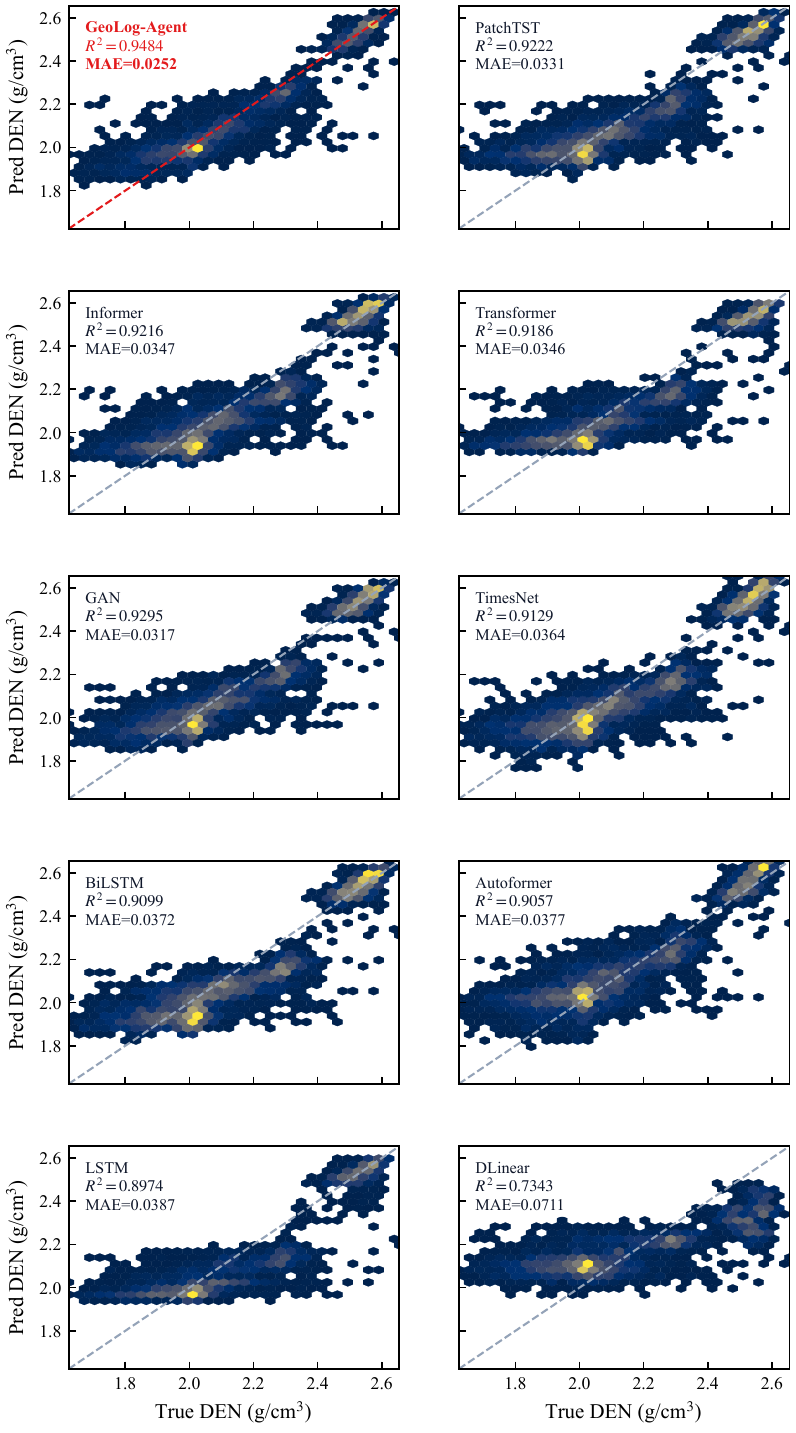}%
\label{fig:scatter_w1_den}}
\caption{True versus predicted hexbin density scatter matrices across all ten models for Well 1 (15-9-15) under M50: (\textbf{a}) AC and (\textbf{b}) DEN.}
\label{fig:scatter_w1}
\end{figure*}

\begin{figure*}[!htbp]
\centering
\subfloat[Well 2 AC hexbin scatter matrix]{\includegraphics[width=0.48\textwidth]{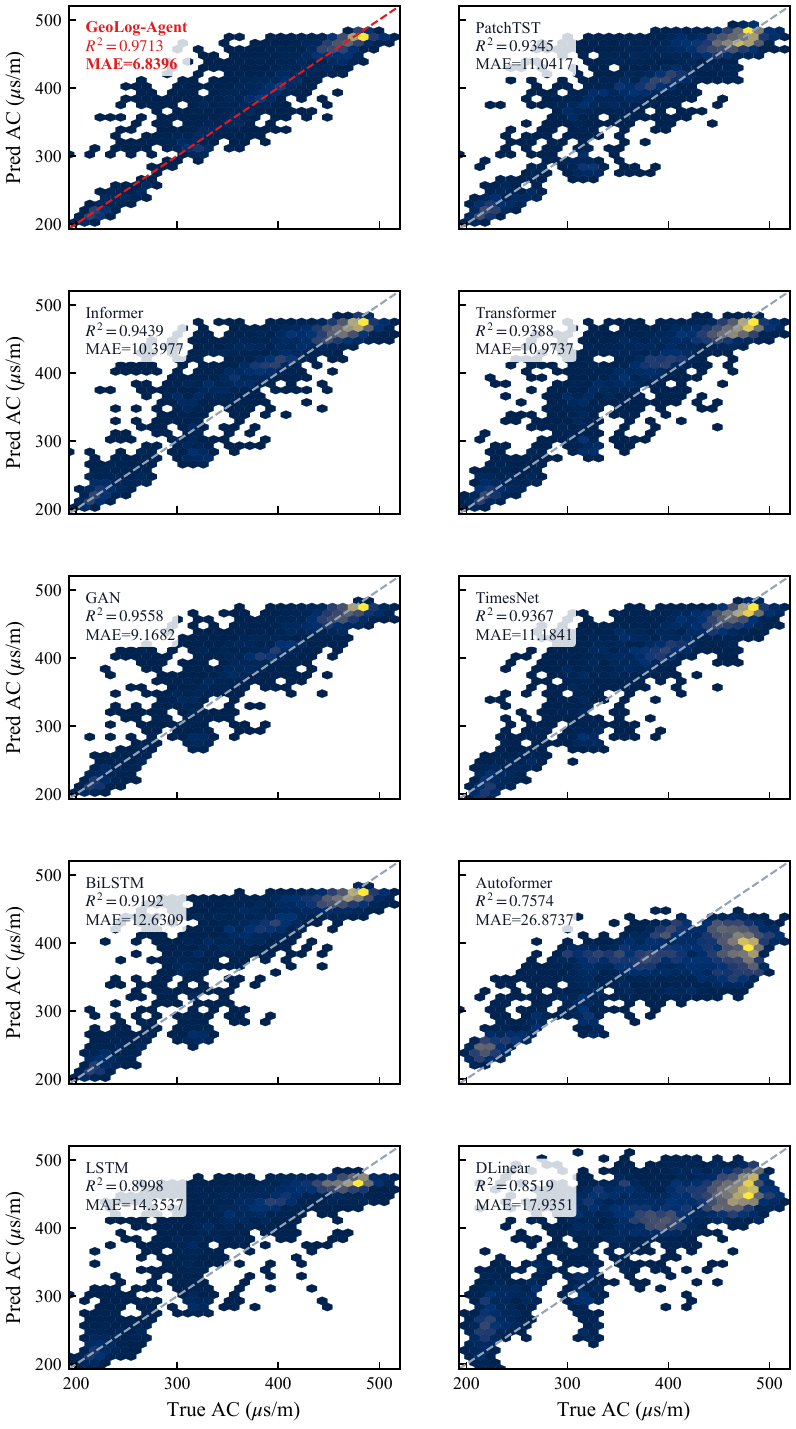}%
\label{fig:scatter_w2_ac}}
\hfil
\subfloat[Well 2 DEN hexbin scatter matrix]{\includegraphics[width=0.48\textwidth]{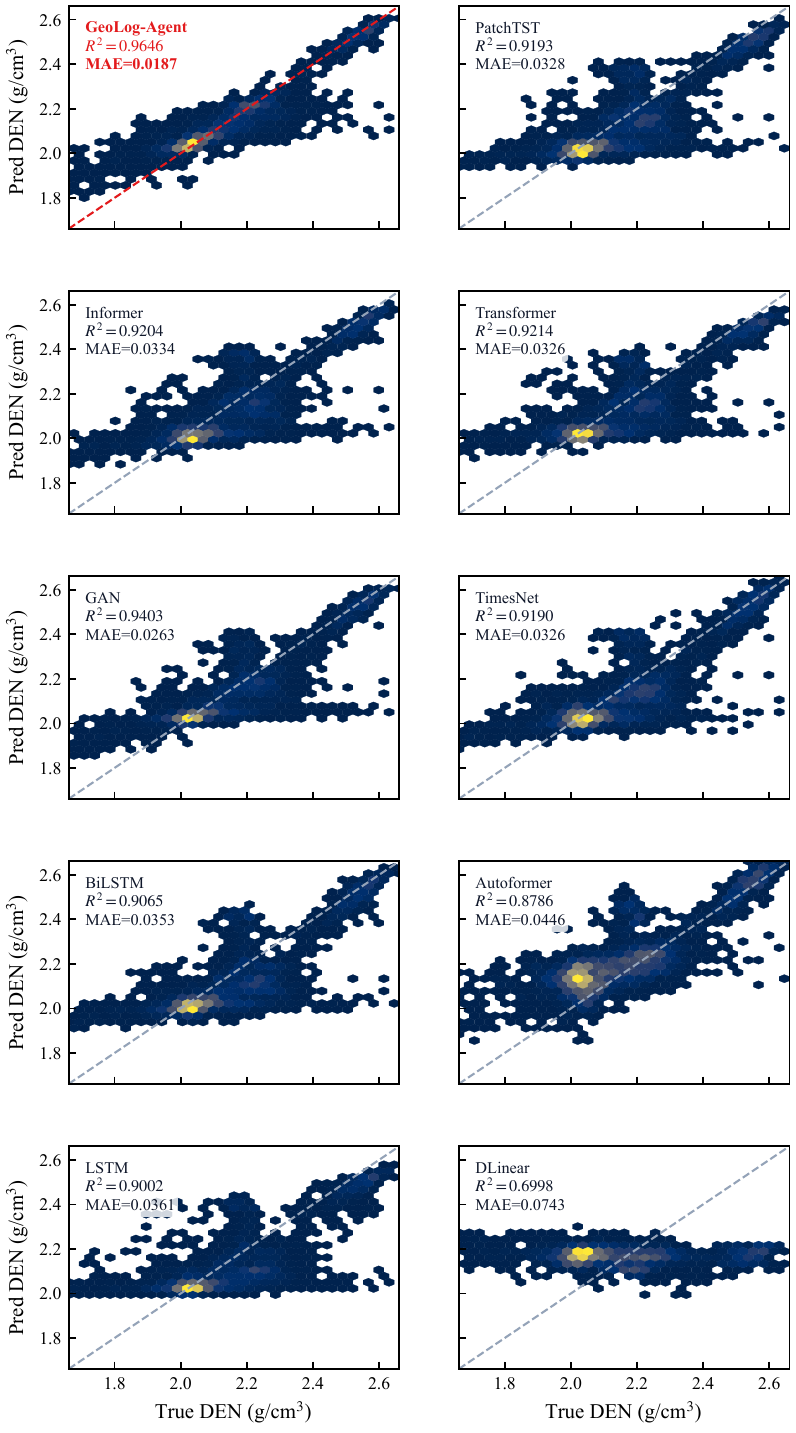}%
\label{fig:scatter_w2_den}}
\caption{True versus predicted hexbin density scatter matrices across all ten models for Well 2 (15-9-17) under M50: (\textbf{a}) AC and (\textbf{b}) DEN.}
\label{fig:scatter_w2}
\end{figure*}

\begin{figure*}[!htbp]
\centering
\subfloat[Per-scenario model ranking matrix]{\includegraphics[width=0.47\textwidth]{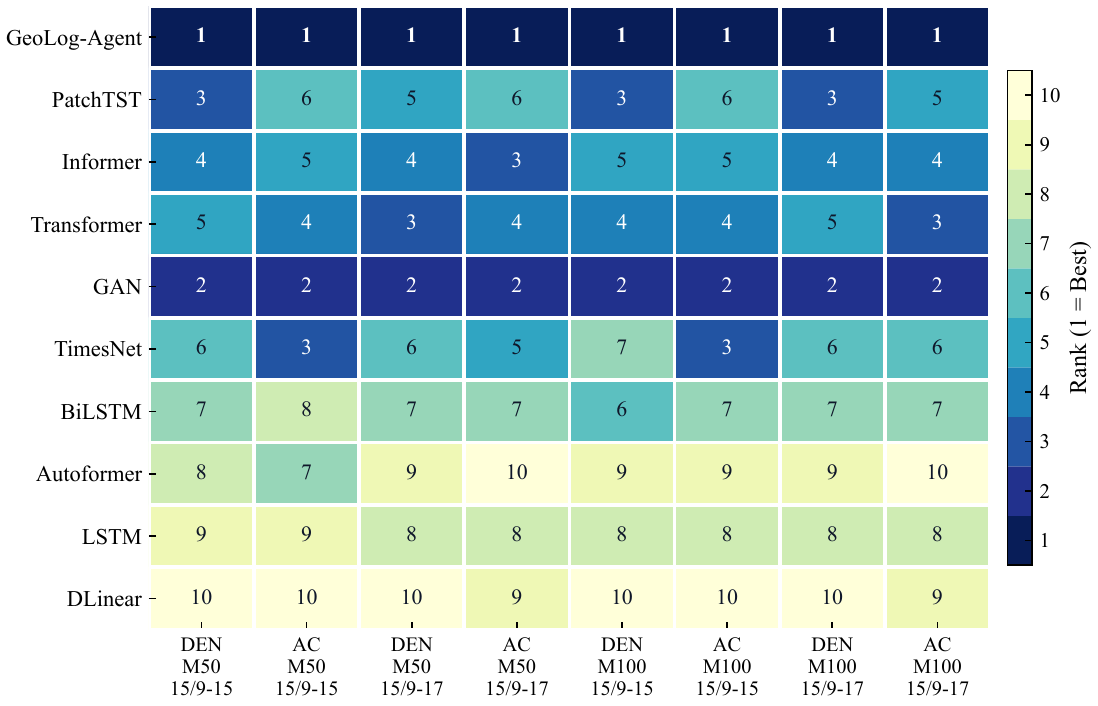}%
\label{fig:rank_matrix}}
\hfil
\subfloat[Depth-sequence DEN error heatmap]{\includegraphics[width=0.49\textwidth]{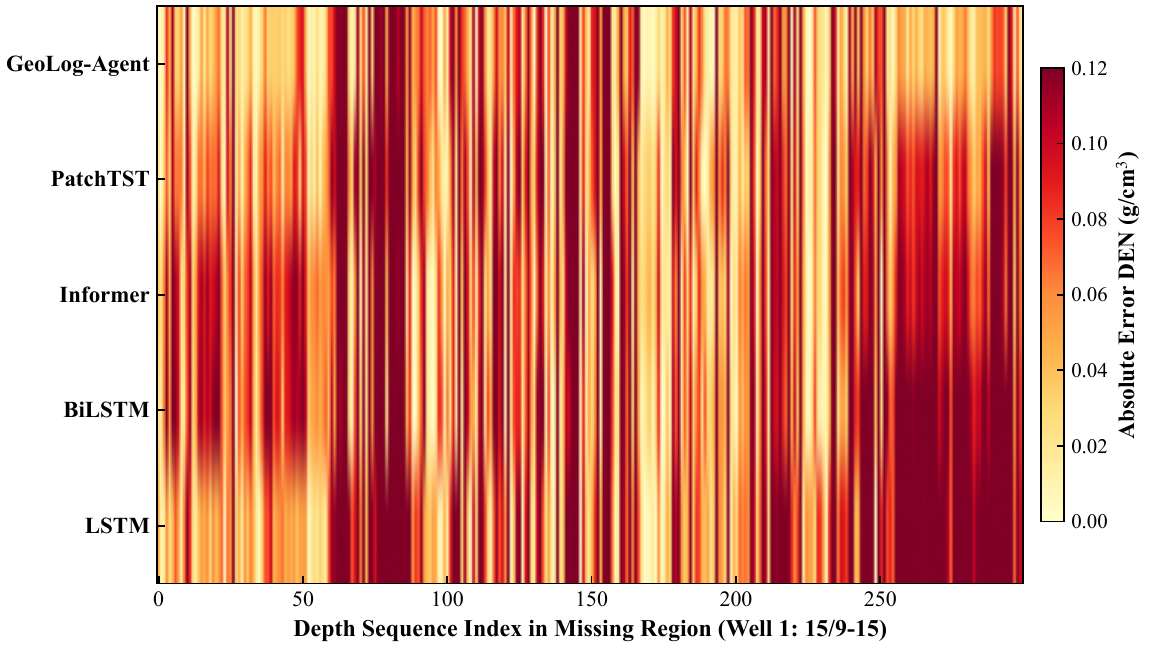}%
\label{fig:err_heatmap_w1}}
\\[4pt]
\subfloat[Boundary continuity across the lower M50 boundary]{\includegraphics[width=0.66\textwidth]{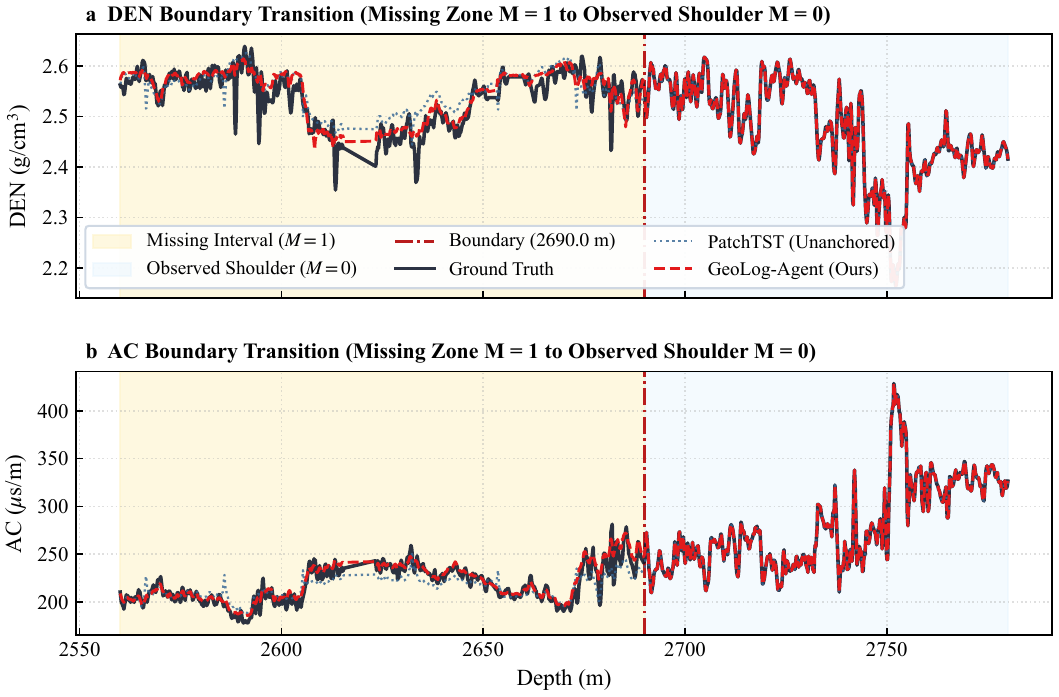}%
\label{fig:fusion}}
\caption{Supplementary error and boundary diagnostics: (\textbf{a}) per-scenario model ranking matrix, (\textbf{b}) depth-sequence DEN absolute error heatmap in Well 1 (15-9-15), and (\textbf{c}) boundary transition across $2690.0\text{ m}$.}
\label{fig:error_heatmap}
\end{figure*}

\begin{figure*}[!htbp]
\centering
\subfloat[Six-axis radar profile under M50]{\includegraphics[width=0.47\textwidth]{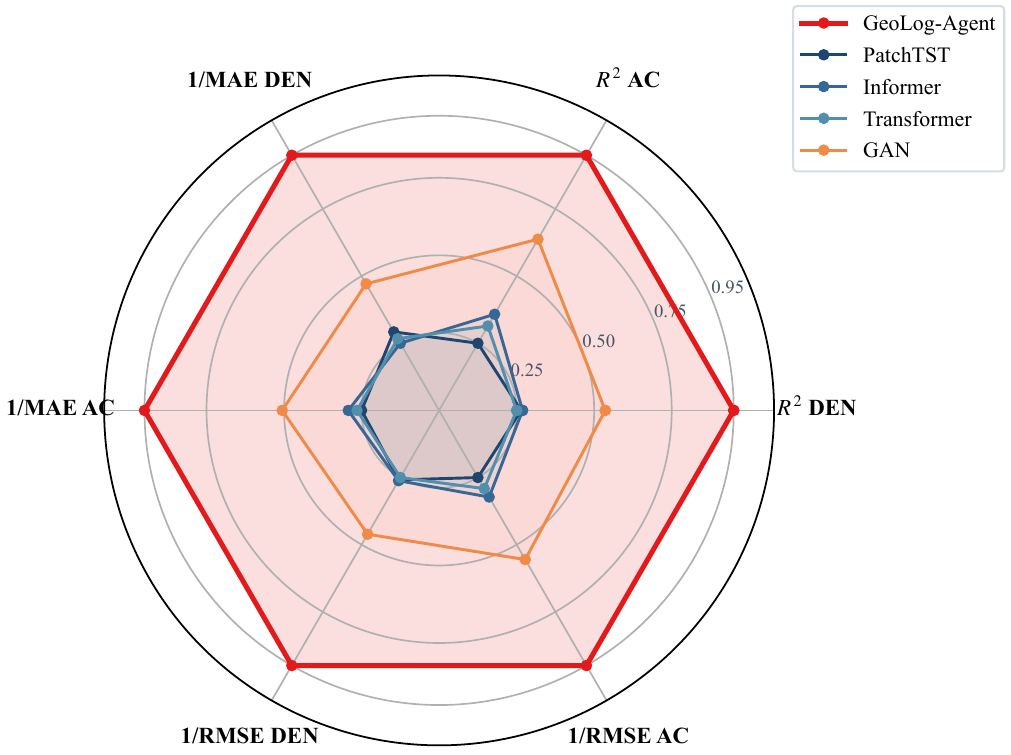}%
\label{fig:radar_m50}}
\hfil
\subfloat[Six-axis radar profile under M100]{\includegraphics[width=0.47\textwidth]{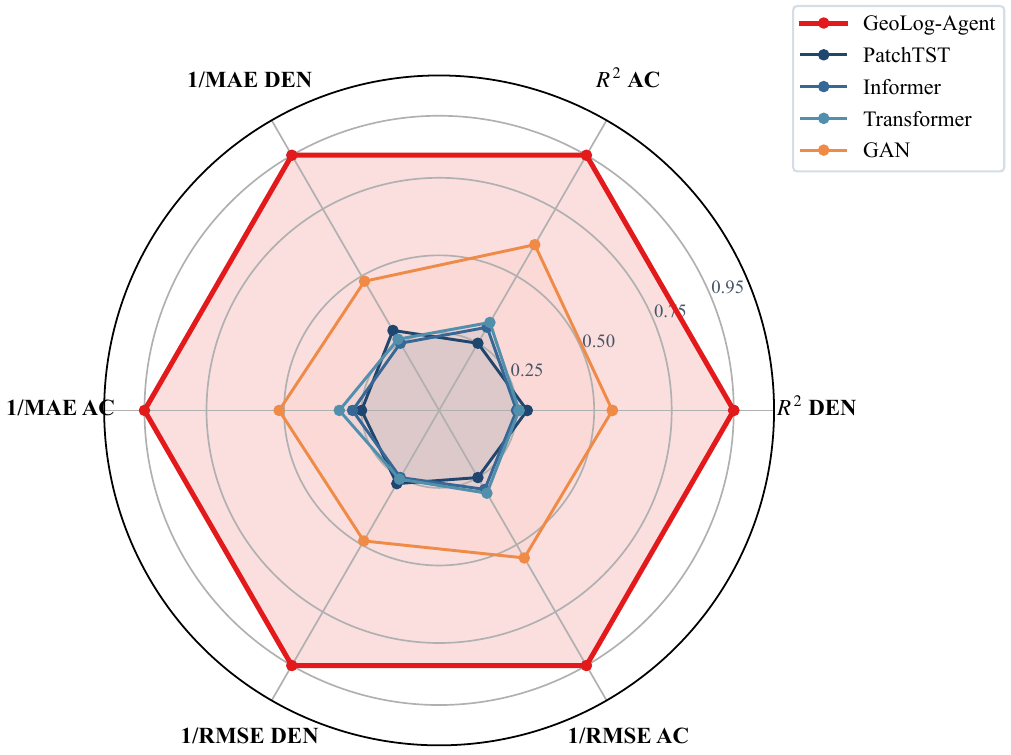}%
\label{fig:radar_m100}}
\caption{Normalized six-axis radar comparisons of the top five models under (\textbf{a}) M50 and (\textbf{b}) M100 protocols.}
\label{fig:radar_appendix}
\end{figure*}

This appendix provides the consolidated per-well and per-curve comparative benchmark breakdown across blind test Well 1 (15-9-15) and Well 2 (15-9-17) under the M50 and M100 protocols (Table~\ref{tab:per_well_breakdown}), the downstream petrophysical inversion and reservoir characterization evaluation (Table~\ref{tab:downstream_app}), and supplementary visual diagnostics (Figs.~\ref{fig:bar_comparisons}--\ref{fig:radar_appendix}).

\begin{table*}[!htbp]
\centering
\caption{Consolidated per-well and per-curve reconstruction performance for bulk density (DEN, $\text{g/cm}^3$) and compressional acoustic transit time (AC, $\mu\text{s/m}$) under the M50 (50\% contiguous missing, evaluated on the full-well reconstructed profile $\hat{\mathbf{Y}}_{\mathrm{full}}$ after observation-preserving fusion) and M100 (100\% full-well blind) protocols across blind test Well 1 (15-9-15) and Well 2 (15-9-17). Bold values indicate the best performance.}
\label{tab:per_well_breakdown}
\footnotesize
\setlength{\tabcolsep}{2.4pt}
\renewcommand{\arraystretch}{1.12}
\resizebox{\textwidth}{!}{%
\begin{tabular}{llcccccccccccc}
\toprule
\multirow{3}{*}{\textbf{Protocol}} & \multirow{3}{*}{\textbf{Model}} & \multicolumn{6}{c}{\textbf{Bulk Density (DEN, $\text{g/cm}^3$)}} & \multicolumn{6}{c}{\textbf{Compressional Transit Time (AC, $\mu\text{s/m}$)}} \\
\cmidrule(lr){3-8} \cmidrule(lr){9-14}
 & & \multicolumn{3}{c}{\textbf{Well 1 (15-9-15)}} & \multicolumn{3}{c}{\textbf{Well 2 (15-9-17)}} & \multicolumn{3}{c}{\textbf{Well 1 (15-9-15)}} & \multicolumn{3}{c}{\textbf{Well 2 (15-9-17)}} \\
\cmidrule(lr){3-5} \cmidrule(lr){6-8} \cmidrule(lr){9-11} \cmidrule(lr){12-14}
 & & $\mathbf{R^2} \uparrow$ & \textbf{MAE} $\downarrow$ & \textbf{RMSE} $\downarrow$ & $\mathbf{R^2} \uparrow$ & \textbf{MAE} $\downarrow$ & \textbf{RMSE} $\downarrow$ & $\mathbf{R^2} \uparrow$ & \textbf{MAE} $\downarrow$ & \textbf{RMSE} $\downarrow$ & $\mathbf{R^2} \uparrow$ & \textbf{MAE} $\downarrow$ & \textbf{RMSE} $\downarrow$ \\
\midrule
\multirow{10}{*}{\textbf{M50}}
 & DLinear     & 0.7343 & 0.0711 & 0.1215 & 0.6998 & 0.0743 & 0.1293 & 0.8712 & 20.9275 & 42.4008 & 0.8519 & 17.9351 & 35.1579 \\
 & Autoformer  & 0.9057 & 0.0377 & 0.0724 & 0.8786 & 0.0446 & 0.0822 & 0.8936 & 17.8242 & 38.5345 & 0.7574 & 26.8737 & 44.9962 \\
 & LSTM        & 0.8974 & 0.0387 & 0.0755 & 0.9002 & 0.0361 & 0.0745 & 0.8846 & 18.7295 & 40.1211 & 0.8998 & 14.3537 & 28.9255 \\
 & BiLSTM      & 0.9099 & 0.0372 & 0.0708 & 0.9065 & 0.0353 & 0.0722 & 0.8872 & 18.8382 & 39.6770 & 0.9192 & 12.6309 & 25.9700 \\
 & TimesNet    & 0.9129 & 0.0364 & 0.0696 & 0.9190 & 0.0326 & 0.0671 & 0.8979 & 17.1196 & 37.7526 & 0.9367 & 11.1841 & 22.9775 \\
 & PatchTST    & 0.9222 & 0.0331 & 0.0658 & 0.9193 & 0.0328 & 0.0670 & 0.8950 & 17.5792 & 38.2676 & 0.9345 & 11.0417 & 23.3754 \\
 & Transformer & 0.9186 & 0.0346 & 0.0673 & 0.9214 & 0.0326 & 0.0661 & 0.8979 & 17.3672 & 37.7504 & 0.9388 & 10.9737 & 22.6033 \\
 & Informer    & 0.9216 & 0.0347 & 0.0660 & 0.9204 & 0.0334 & 0.0666 & 0.8977 & 17.4134 & 37.7761 & 0.9439 & 10.3977 & 21.6290 \\
 & GAN         & 0.9295 & 0.0317 & 0.0626 & 0.9403 & 0.0263 & 0.0577 & 0.9171 & 15.2138 & 34.0106 & 0.9558 &  9.1682 & 19.2081 \\
 & \textbf{GeoLog-Agent} & \textbf{0.9484} & \textbf{0.0252} & \textbf{0.0536} & \textbf{0.9646} & \textbf{0.0187} & \textbf{0.0444} & \textbf{0.9366} & \textbf{12.5583} & \textbf{29.7437} & \textbf{0.9713} & \textbf{6.8396} & \textbf{15.4857} \\
\midrule
\multirow{10}{*}{\textbf{M100}}
 & DLinear     & 0.5894 & 0.1222 & 0.1511 & 0.2846 & 0.1691 & 0.1996 & 0.6485 & 51.1896 & 70.0372 & 0.7065 & 39.1724 & 49.4920 \\
 & Autoformer  & 0.8150 & 0.0778 & 0.1014 & 0.5253 & 0.1319 & 0.1626 & 0.7047 & 40.8465 & 64.1946 & 0.4940 & 56.5439 & 64.9855 \\
 & LSTM        & 0.8412 & 0.0699 & 0.0940 & 0.7813 & 0.0824 & 0.1104 & 0.7443 & 33.1879 & 59.7346 & 0.8549 & 23.3868 & 34.7946 \\
 & BiLSTM      & 0.8519 & 0.0696 & 0.0907 & 0.8192 & 0.0719 & 0.1003 & 0.7495 & 33.7133 & 59.1266 & 0.8673 & 22.8964 & 33.2767 \\
 & TimesNet    & 0.8414 & 0.0697 & 0.0939 & 0.8200 & 0.0716 & 0.1001 & 0.7665 & 30.8266 & 57.0770 & 0.8717 & 24.4117 & 32.7230 \\
 & PatchTST    & 0.8700 & 0.0630 & 0.0850 & 0.8381 & 0.0679 & 0.0950 & 0.7625 & 31.1779 & 57.5629 & 0.8870 & 21.3484 & 30.7049 \\
 & Informer    & 0.8657 & 0.0666 & 0.0864 & 0.8362 & 0.0689 & 0.0955 & 0.7634 & 31.3956 & 57.4615 & 0.8984 & 20.3115 & 29.1150 \\
 & Transformer & 0.8686 & 0.0641 & 0.0855 & 0.8347 & 0.0699 & 0.0959 & 0.7637 & 31.1617 & 57.4215 & 0.9018 & 19.3958 & 28.6295 \\
 & GAN         & 0.8854 & 0.0585 & 0.0798 & 0.8731 & 0.0573 & 0.0841 & 0.8048 & 28.5244 & 52.1955 & 0.9205 & 17.3793 & 25.7521 \\
 & \textbf{GeoLog-Agent} & \textbf{0.9235} & \textbf{0.0442} & \textbf{0.0652} & \textbf{0.9068} & \textbf{0.0452} & \textbf{0.0720} & \textbf{0.8511} & \textbf{24.2737} & \textbf{45.5878} & \textbf{0.9430} & \textbf{13.7934} & \textbf{21.8111} \\
\bottomrule
\end{tabular}%
}
\end{table*}

\begin{table*}[!htbp]
\centering
\caption{Downstream reservoir characterization and seismic inversion performance evaluated directly on the reconstructed target intervals ($M_{t,j}=1$, averaged across M50 and M100 on blind test Wells 15-9-15 and 15-9-17; effective porosity $\phi_D = (\rho_{\mathrm{ma}} - \rho_b)/(\rho_{\mathrm{ma}} - \rho_{\mathrm{fl}})$ with $\rho_{\mathrm{ma}} = 2.65\text{ g/cm}^3, \rho_{\mathrm{fl}} = 1.00\text{ g/cm}^3$, and four-zone pay delineation defined by standard North Sea shale-volume, porosity, and resistivity cutoffs).}
\label{tab:downstream_app}
\footnotesize
\setlength{\tabcolsep}{3.8pt}
\renewcommand{\arraystretch}{1.15}
\resizebox{\textwidth}{!}{%
\begin{tabular}{lcccccc}
\toprule
\multirow{2}{*}{\textbf{Model}} & \multicolumn{2}{c}{\textbf{Acoustic Impedance $Z_p$ ($\text{MPa}\cdot\text{s/m}$)}} & \multicolumn{2}{c}{\textbf{Reservoir Porosity $\phi_D$ (p.u.)}} & \multicolumn{2}{c}{\textbf{Pay-Zone Delineation}} \\
\cmidrule(lr){2-3} \cmidrule(lr){4-5} \cmidrule(lr){6-7}
 & $\mathbf{R^2} \uparrow$ & \textbf{RMSE} $\downarrow$ & $\mathbf{R^2} \uparrow$ & \textbf{RMSE} $\downarrow$ & \textbf{Macro-F1 (\%)} $\uparrow$ & \textbf{Accuracy (\%)} $\uparrow$ \\
\midrule
DLinear     & 0.5904 & 1.7655 & 0.3710 & 9.9259 & 38.39 & 71.99 \\
Autoformer  & 0.6134 & 1.7342 & 0.7240 & 6.4740 & 57.38 & 78.68 \\
LSTM        & 0.7654 & 1.3422 & 0.7998 & 5.6444 & 56.94 & 84.55 \\
BiLSTM      & 0.7743 & 1.3072 & 0.8115 & 5.4497 & 58.19 & 85.20 \\
TimesNet    & 0.7841 & 1.2753 & 0.8287 & 5.2316 & 59.44 & 85.67 \\
PatchTST    & 0.7961 & 1.2391 & 0.8441 & 4.9367 & 59.40 & 85.40 \\
Informer    & 0.7982 & 1.2313 & 0.8363 & 5.0929 & 61.15 & 86.21 \\
Transformer & 0.7995 & 1.2310 & 0.8427 & 5.0019 & 61.10 & 86.23 \\
GAN         & 0.8209 & 1.1514 & 0.8749 & 4.4871 & 64.36 & 87.46 \\
\textbf{GeoLog-Agent} & \textbf{0.8501} & \textbf{1.0349} & \textbf{0.9216} & \textbf{3.5895} & \textbf{76.34} & \textbf{91.25} \\
\bottomrule
\end{tabular}%
}
\end{table*}

\section*{Funding}
This work was supported by the Major Scientific and Technological Projects of CNPC (Grant No. ZD2019-183-004), and the Fundamental Research Funds for the Central Universities (Grant No. 20CX05019A).

\section*{Data and Code Availability}
The Norwegian North Sea well-log and lithofacies dataset analyzed in this study is publicly available from the FORCE 2020 Machine Learning Competition repository released by the Norwegian Offshore Directorate \cite{bormann2020force}. All deep learning models were implemented in PyTorch and executed on an NVIDIA RTX GPU. No datasets, per-sample predictions, model checkpoints, or data attachments are distributed with this preprint.

\section*{Declaration of Competing Interest}
The authors declare that they have no known competing financial interests or personal relationships that could have appeared to influence the work reported in this paper.

\section*{CRediT Authorship Contribution Statement}
\textbf{Yongan Zhang:} Methodology, Software, Validation, Formal analysis, Writing -- original draft, Visualization. \textbf{Yitao Yu:} Data curation, Investigation, Validation, Visualization. \textbf{Yizheng Li:} Software, Investigation, Data curation. \textbf{An Gong:} Funding acquisition. \textbf{Youzhuang Sun:} Conceptualization, Supervision, Project administration, Writing -- review \& editing.

\end{document}